%% file: domain_transform_solver.tex
\pdfoutput=1
\documentclass[runningheads]{llncs}
\usepackage{graphicx}
\usepackage{amsmath,amssymb} 

\usepackage{color}
\usepackage[width=122mm,left=12mm,paperwidth=146mm,height=193mm,top=12mm,paperheight=217mm]{geometry}

\newcommand{\eg}{\textit{e}.\textit{g}.,~}
\newcommand{\etal}{\textit{et} \textit{al}.}
\usepackage{hyperref}
\usepackage{subcaption}
\usepackage{stackengine}
\usepackage{multirow}
\usepackage{tabularx}
\begin{document}
\pagestyle{headings}
\mainmatter
\def\ECCV18SubNumber{1189}  

\title{The Domain Transform Solver} 

\titlerunning{The Domain Transform Solver}

\authorrunning{Akash Bapat and Jan-Michael Frahm.}

\author{Akash Bapat and Jan-Michael Frahm\\
\institute{Department of Computer Science,\\
The University of North Carolina at  Chapel Hill}
%
{\tt\small \{akash,jmf\}@cs.unc.edu}
}
\maketitle

\captionsetup[subfigure]{font=scriptsize,labelfont=scriptsize}
\begin{abstract}
We present a framework for edge-aware optimization that is an order of magnitude faster than the state of the art while having comparable performance.
Our key insight is that the optimization can be formulated by leveraging properties of the domain transform~\cite{gastal2011domain}, a method for edge-aware filtering that defines a distance-preserving 1D mapping of the input space.
This enables our method to improve performance for a variety of problems including stereo, depth super-resolution, and render from defocus, while keeping the computational complexity linear in the number of pixels.
Our method is highly parallelizable and adaptable, and it has demonstrable scalability with respect to image resolution.
\end{abstract}
\input{intro}
\input{related_work}
\input{approach}
\input{results}
\input{conclusion}
\bibliographystyle{splncs}
\bibliography{domain_transform_references}
\end{document}

%% file: intro.tex
\section{Introduction}
Edge-aware optimization is a widely utilized tool in computer vision.
It is applied to a large variety of tasks, including semantic segmentation~\cite{krahenbuhl2011efficient}, stereo~\cite{bleyer2011patchmatch}, recoloration \cite{levin2004colorization}, and optical flow~\cite{revaud2015epicflow}.
This has been motivated by the intuition that similar-looking pixels should have similar properties.
For this reason, a wide variety of edge-aware filtering algorithms have been developed, including the bilateral filter~\cite{tomasi1998bilateral}, anisotropic diffusion~\cite{perona1990scale}, and edge-avoiding wavelets~\cite{fattal2009edge}, all of which identify similar-looking pixels.
However, using such filters in optimization frameworks typically leads to slow algorithms, and while high-level groupings like super-pixels can be used to compensate for this sluggishness~\cite{lu2013patch}, the color-space clusterings of such approaches are not guaranteed to respect the semantics of the underlying domain, which often leads to processing artifacts. 

\begin{figure}
    \centering
    \captionsetup[subfigure]{font=scriptsize,labelfont=scriptsize}
    \begin{subfigure}[t]{0.195\textwidth}
        \centering
        \includegraphics[width=\textwidth] {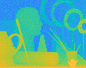}
        \caption{\centering Low-res. depthmap}
    \end{subfigure}%
    ~\hspace{-1mm}
      \begin{subfigure}[t]{0.195\textwidth}
        \centering
        \includegraphics[width=\textwidth] {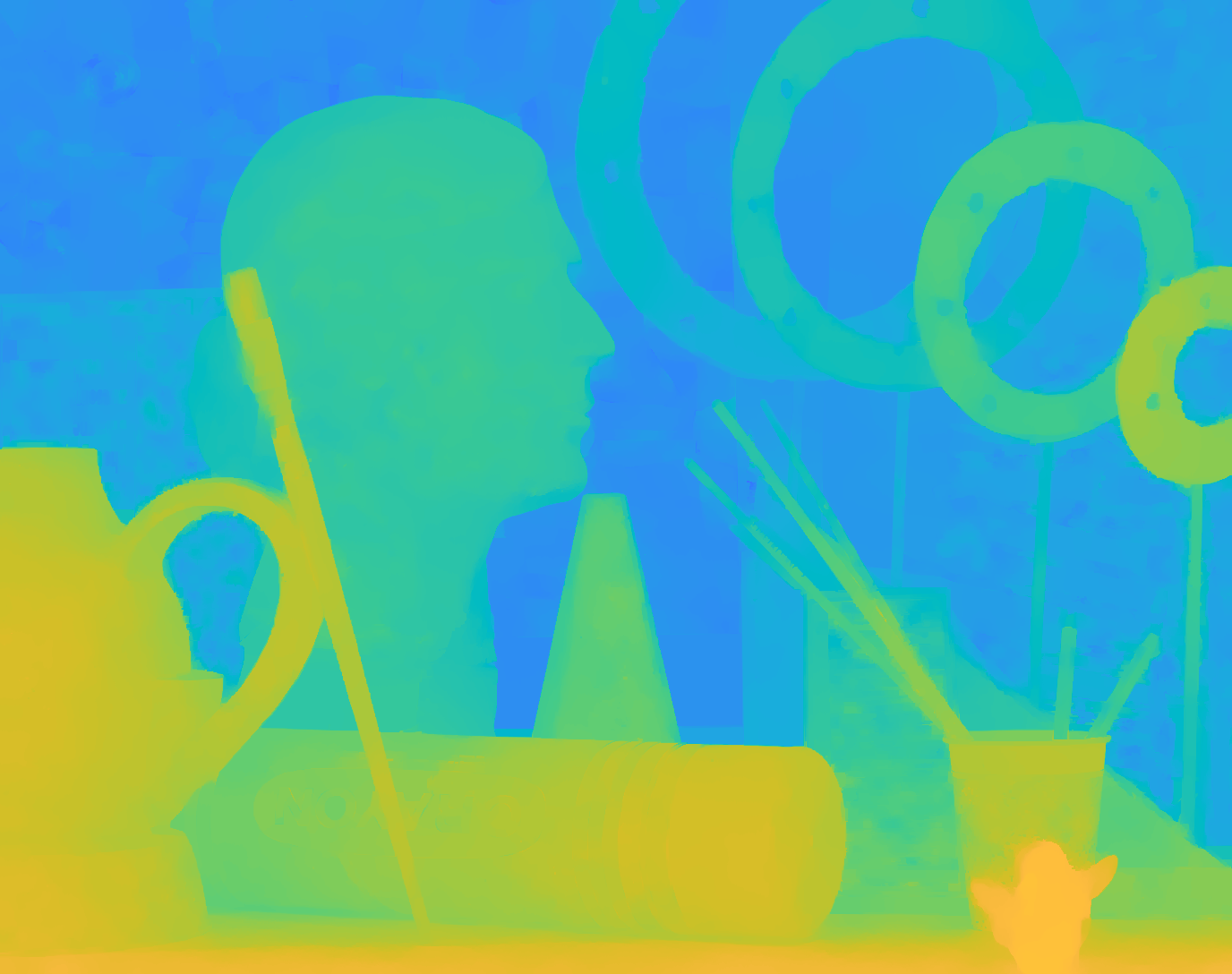}
        \caption{\centering Our 16x upsampled result}
    \end{subfigure}%
    ~\hspace{-1mm}
     \centering
    \begin{subfigure}[t]{0.195\textwidth}
        \centering
        \includegraphics[trim={20cm 26cm 15cm 0cm},clip,width=\textwidth] {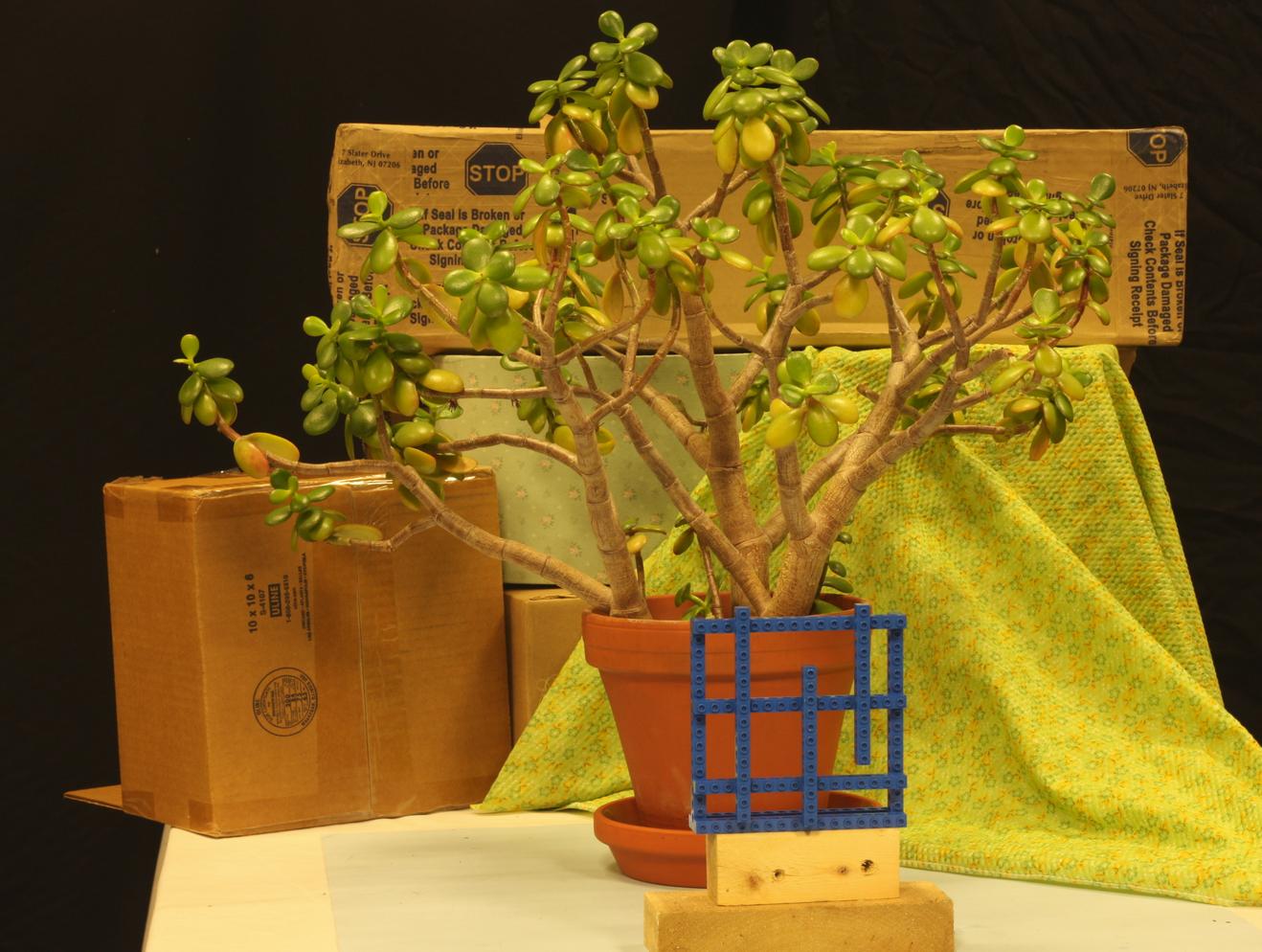}
        \caption{\centering Color image}
    \end{subfigure}%
    ~\hspace{-1mm}
    \begin{subfigure}[t]{0.195\textwidth}
        \centering
        \includegraphics[trim={20cm 26cm 15cm 0cm},clip,width=\textwidth]{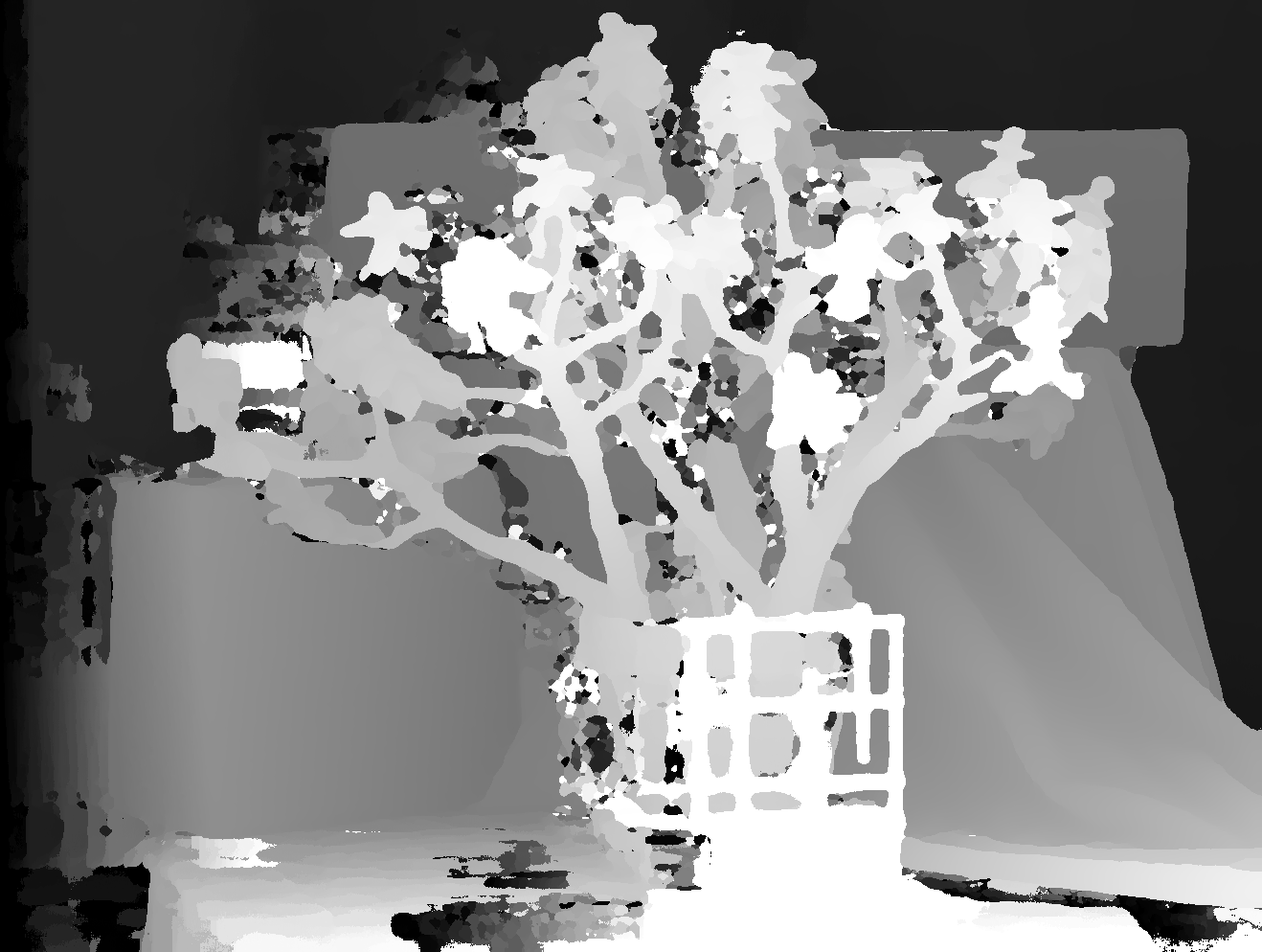}
        \caption{\centering Target image}
    \end{subfigure}%
    ~\hspace{-1mm}
        \begin{subfigure}[t]{0.195\textwidth}
        \centering
        \includegraphics[trim={20cm 26cm 15cm 0cm},clip,width=\textwidth]{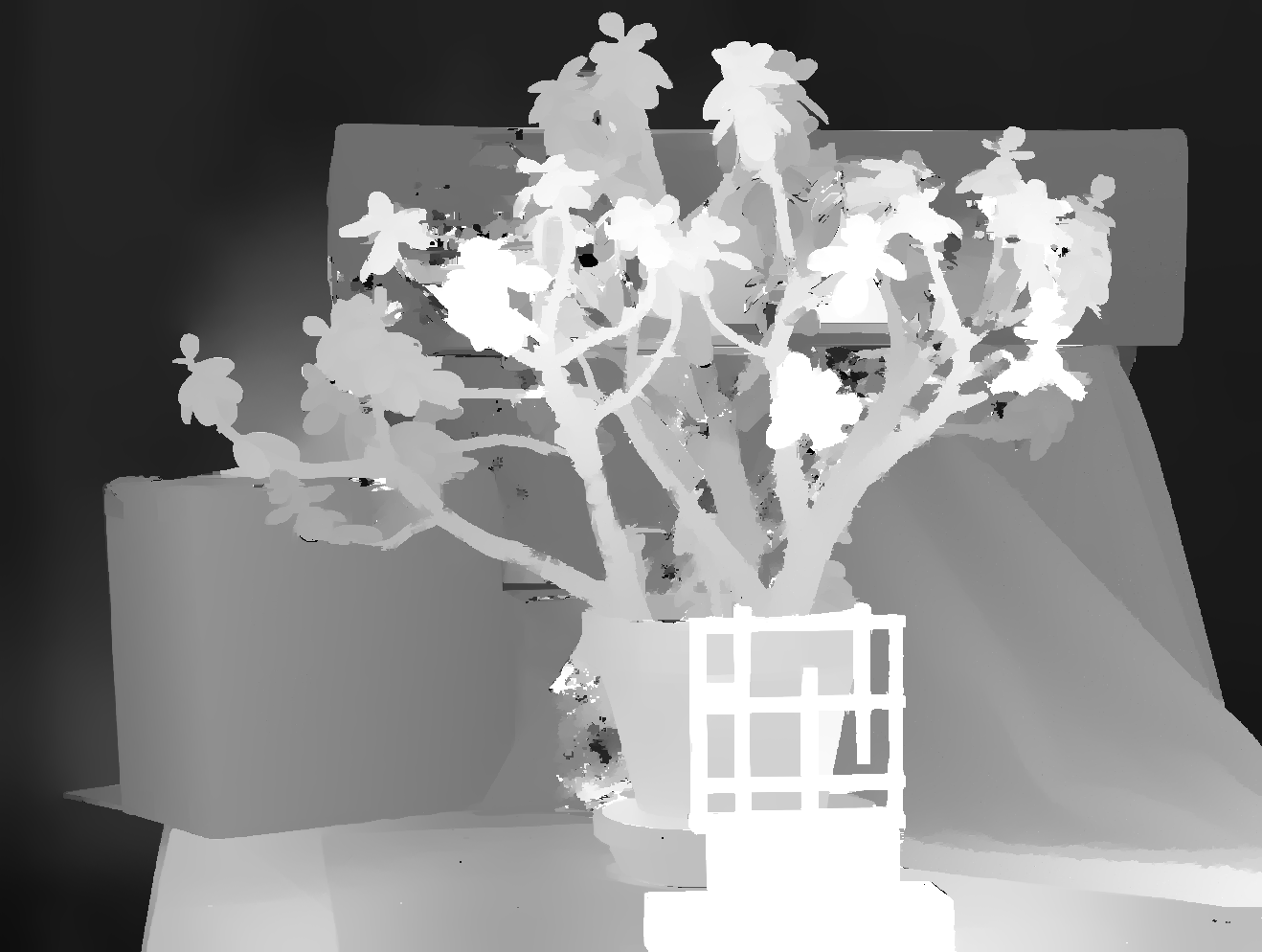}
        \caption{\centering Our result}
    \end{subfigure}%
    \caption{\label{fig:intro_poster}Our domain transform solver can tackle a variety of problems.
(a,b) shows our result for depth super-resolution using a high resolution color image as a reference.
(c,d) shows a color image from the Middlebury dataset~\cite{scharstein2014high} with an initialization obtained from MC-CNN~\cite{zbontar2016stereo}, which is then refined to obtain our result (e) that is blurred in an edge-aware sense.}
\end{figure}

We propose a general optimization framework that directly operates in the pixel space while maintaining distances in the combined color and pixel space with an edge-aware regularizer.
The framework can be applied for a variety of optimization problems, as we demonstrate in Fig.~\ref{fig:intro_poster} and Sec.~\ref{sec:approach}.
Our method achieves competitive performance in applications like stereo optimization (Sec.\ref{sec:stereo_optimization}), rendering from defocus (Sec.\ref{sec:synthetic_defocus}), and depth super-resolution (Sec.\ref{sec:depth_super_resolution}).
This advantage becomes more pronounced with increasing image resolution, as well as a growing number of image channels.
At the same time, our approach is independent of blur kernel sizes, which is not the case for existing bilateral solvers.
This becomes crucial for applications where the data is of high resolution and high dimensionality, for example satellite imagery where a single image has typical resolutions of more than 100 million pixels with as many as 16 spectral bands.

The remainder of the paper is organized as follows: Sec.~\ref{sec:related_work} describes the traditional approaches the computer vision community has developed for edge-aware filtering and optimization. Sec.~\ref{sec:approach} derives our domain transform solver (DTS) optimization framework and highlights its similarities as well as dissimilarities with previous work.
We also describe how our framework can be adapted for various vision tasks. 
In Sec.~\ref{sec:results}, we provide quantitative evaluation as well as validation of the timing performance.
Finally, we conclude in Sec.~\ref{sec:conclusion} and provide some future directions for expanding on our approach.

%% file: related_work.tex
\section{Related Work}
\label{sec:related_work}
Next, we briefly review the most relevant prior work related to edge-based filtering and optimization, namely: implementations for bilateral filters, optimizations leveraging superpixel segmentation, machine learning for edge-aware filtering, the domain transform and its filtering applications, and bilateral solvers.

\paragraph{Bilateral filters}
The bilateral filter was introduced by Tomasi and Manduchi~\cite{tomasi1998bilateral} and presented one of the initial edge-aware blurring techniques. The major bottleneck to bilateral filtering is that it is costly to compute, especially at large blur windows.
Since its invention, there have been multiple approaches introduced to speed up the bilateral filter~\cite{weiss2006fast}, \cite{adams2010fast}, \cite{chen2007real}, \cite{yang2015constant}, \cite{elad2002origin}: Durand and Dorsey approximate the bilateral filter by a piece-wise linear approximation~\cite{durand2002fast}.
Pham and van Vliet proposed to approximate the bilateral filter using two 1-D bilateral kernels~\cite{pham2005separable}.
Paris and Durand proposed to treat the image as a 5-D function of color and pixel space and then apply 1-D blur kernels in this high dimensional space\cite{paris2006fast}.
These approximations to the bilateral filter, or the use of 1-D kernels in higher-dimensional spaces, enables decoupling the 2-D adaptive bilateral kernel into 1-D kernel, reducing the computational cost significantly.
When used as a post-processing step, the bilateral filter removes noise in homogeneous regions but is sensitive to artifacts such as salt and pepper noise~\cite{zhang2008multiresolution}.
Our method emphasizes edge-aware concepts in the same spirit as bilateral filters, but our formulation is fundamentally different in that it provides a generalized framework for domain transform optimization.
\paragraph{Superpixels}
To combat issues of computational complexity during bilateral optimizations, several approaches leverage superpixels, \textit{i.e.}, they group pixels together based on appearance and location.
Superpixel extraction algorithms like SLIC~\cite{achanta2012slic} are often used in optimization problems for two major reasons: 1) They reduce the number of variables in the optimization, and 2) they adhere to color and (implicitly) object boundaries.
In one application, B{\'o}dis-Szomor{\'u}~\etal~\cite{bodis2015superpixel} use sparse Structure-from-Motion(SfM) data, image gradients, and superpixels for surface reconstruction to ensure that the edges of triangles are aligned to the edges in the image.
Lu~\etal~\cite{lu2013patch} use SLIC superpixels to enforce spatially consistent depths in a PatchMatch-based matching framework~\cite{barnes2009patchmatch} to estimate stereo.
Using superpixels inherently assumes local consistency and perfect segmentation, which often does not hold in practice.
For example, in stereo algorithms, superpixels may cover regions with similar color but drastically different depths.
At such regions, the pixels in a superpixel are incorrectly grouped because they have different depths.
Although algorithm parameters can be tuned, the trade-off in coherence versus conciseness is a limiting factor in the utility of superpixel approaches.
%
\paragraph{Machine learning for edge-awareness}
Yan~\etal~\cite{yang2010svm} used support vector machines (SVM) to mimic a bilateral filter by using the exponential of spatial and color distances as feature vectors to represent each pixel.
Traditionally, conditional random fields (CRFs) are used for enforcing pair-wise pixel smoothness via the Potts potential.
For example, Kr{\"a}henb{\"u}hl and Koltun~\cite{krahenbuhl2011efficient} proposed to use the permutohedral lattice data structure~\cite{adams2010fast}, which is typically used in fast bilateral implementations, to accelerate inference in a fully connected CRF by using Gaussian distances in space and color. 
With the explosion of compute capacity and convolutional models in the vision community, there are also deep-learning methods that attempt to achieve edge-aware filtering.
Chen~\etal~\cite{chen2016deeplab} presented DeepLab to perform semantic segmentation; there, they use the fully connected CRF from Kr{\"a}henb{\"u}hl and Koltun~\cite{krahenbuhl2011efficient} on top of their convolutional neural network (CNN) to improve the localization of object boundaries, which typically suffers in a CNN setting due multiple max-poolings and the use of low-resolution images.
Xu~\etal~\cite{xu2015deep} presented a framework to learn edge-aware operators from the data to mimic various traditional handcrafted filters like the bilateral, weighted median, and weighted least squares filters~\cite{farbman2008edge}.
However, machine learning approaches require large amounts of training data specific to a task, as well as significant compute power, while our approach works without any task-dependent training and runs efficiently on a single GPU.
\paragraph{The domain transform}
Gastal and Oliveira~\cite{gastal2011domain} introduced the domain transform, a novel and efficient method for edge-aware filtering that is akin to bilateral filters.
The domain transform is defined as a 1-D isometric transformation of a multi-valued 1-D function such that the distances in the range and domain are preserved.
(See Sec.~\ref{sec:domain_transform_derivation} for more details.)
When applied to a 1-D image with multiple color channels, the transformation maps the distances in color and pixel space into a 1-D distance in the transformed space.
When the \textit{scalar} distance is measured in the transformed space, it is equivalent to measuring the \textit{vector} distance in [R,G,B,X] space. 
This has the benefit of dimensionality reduction, leading to a fast edge-aware filtering technique which respects edges in color while blurring nearby pixels. 
To apply the domain transform to a 2-D image, the authors apply two passes, once in the X direction and once in Y.
Applying the domain transform to an image results in a filtering effect, while in our case we optimize according to an objective function.
Chen~\etal~\cite{chen2016semantic} proposed to perform edge-aware semantic segmentation using deep learning and use the domain transform filter in their end-to-end training of their deep-learning framework.
They also alter the definition of what is considered as `edge' by learning an edge prediction network, and they then use the learned edge-map in a domain transform.
Their application of the domain transform is in the form of a \textit{filter}, and hence is similar to one iteration of our method.
We use the domain transform in our method in an iterative fashion in an optimization framework because it provides an efficient way to compute the local edge-aware mean.
\paragraph{Bilateral solvers}
More recently, Barron~\etal~\cite{barron2015fast} suggested to view a color image as a function of the 5-D space [Y,U,V,X,Y], which they call the `bilateral space' to estimate stereo for rendering defocus blur.
They proposed to transform the stereo optimization problem by expressing the problem variables in the bilateral space and optimize in this new space. We will refer to this method as BL-Stereo. 
Barron and Poole's\cite{barron2016fast} Bilateral Solver (BL-Solver), on the other hand, solves a linear optimization problem in the bilateral space, which is different from BL-Stereo.
In this setting, they require a target map to enhance, as well as a confidence map for the target map.
The linearization of the problem allows them to converge to the solution faster. (See Sec~\ref{sec:optimization_framework} for more details.)
Both of these approaches quantize the 5-D space into a grid, where the grid size is governed by the blur kernel sizes.
This reduces the number of optimization variables and hence the complexity, leading to fast runtimes.
 
Our work is closely related to Barron~\etal~\cite{barron2015fast} and Barron and Poole~\cite{barron2016fast} in that we are targeting the same goal of developing general solvers that are edge-aware and fast. 
The gridding strategy of these previous methods scales well with higher blur amounts/windows.
However, using higher blur windows is not a scalable option as image resolution increases, especially in large-resolution imagery where it is important to maintain fine details, such as multi-camera capture for virtual reality and satellite imagery.
In contrast, our method does not require large blur kernels to be efficient.
Our method operates on the pixels themselves, and hence inherently has a large number of optimization variables.
Despite this, our approach is inherently parallelizable, making it easy to implement it on GPUs.
Our method does not depend on the blur kernel size, and hence scales well with higher image resolutions with large and small blur kernels.
 
In the following sections, we present our general optimization framework and demonstrate its performance on a variety of problems.
We present quantitative evaluations to show the competitive accuracy of our method, as well as the significant speed-up that it provides.

%% file: approach.tex
\section{Approach}
\label{sec:approach}
Edge-aware filtering techniques like the bilateral filter smooth similar looking regions of the image while preserving crisp edges. This is especially useful for smoothing depthmaps, where we want to preserve sharp discontinuities in depth by not filtering across depth edges while smoothing out planar regions.
Although edge-aware filtering has been used for stereo as a post processing step~\cite{zbontar2016stereo}, as well as during optimization~\cite{bleyer2011patchmatch}, this approach increases the computational complexity of the algorithm substantially due to the data-dependent smoothing kernel.
 We consider an algorithm a \textit{filtering technique} when the filter operates on the input image to produce an output image. On the other hand, we consider an algorithm a \textit{solver}, when it uses one or more input images and optimizes towards a goal defined by a cost/loss function.
 \subsection{Optimization framework}
 \label{sec:optimization_framework}
First, we introduce our efficient domain transform solver (DTS), which leverages an efficient way of expressing distances using isometry.
The DTS solves the following optimization problem:
\begin{equation}
\begin{aligned}
& \underset{z}{\text{min}}~F\left(z\right) = 
& \underset{z}{\text{min}}
& & \dfrac{\lambda}{2} \sum_{i}  \underbrace{\left( z_{i} - \bar{z}_{N} \right)^{2}}_{= e\left( z_{i}, N\right)} + \sum_{i} c_{i} \left( z_{i} - t_{i}\right)^{2} + \dfrac{1}{2}\sum_{m}\lambda_{m}\Phi_{m}\left(z\right).
\end{aligned}
\label{eqn:dts_objective}
\end{equation}
Here, the $z_{i}$ are the values we want to estimate,~\eg disparity and color, at the $i^{th}$ pixel of an image.
The initial target estimate $t_i$ with a confidence $c_i$ are also given for the $i^{th}$ pixel.
This optimization objective has an edge-aware regularizer $e\left( z_{i}, N\right)$, which forces the $z_{i}$  to be similar to the mean of the neighborhood $N$, computed in an edge-aware sense.
Hence, the neighborhood's size changes for each $z_i$ according to the image content.
Intuitively, by forcing $z_{i}$ to be similar to the edge-aware mean $\bar{z}_{N}$, we emulate the bilateral filter's properties so that $z_{i}$ is similar to the other $z_{k}$ which contribute to the mean $\bar{z}_{N}$ only when $z_{i}$ and $z_{k}$ are similar in color and close in pixel distance.
  This edge-aware mean is  $\bar{z}_{N} = \dfrac{\sum_{i,j} W_{i,j}*z_{j}}{\sum_{i,j} W_{i,j}}$, where $W_{i,j}$ takes into account the pixel color similarity as well as pixel distance between pixel $i$ and $j$; see Sec.\ref{sec:domain_transform_derivation} for a derivation of $W_{i,j}$.
  We compute $\bar{z}_{N}$ using the domain transform, which enables us to evaluate our pair-wise regularizer faster than traditional approaches; see Sec.~\ref{sec:domain_transform_derivation} for more details.
$\Phi_{m}\left(z\right)$ is an application-dependent term with a weighting factor of $\lambda_{m}$.
For example, for stereo, $\Phi_{m}\left(z\right)$ could be the photometric matching cost for the left-right image pair.

In all applications, our method aims to solve Eq.(\ref{eqn:dts_objective}). The minimum at the point of the solution necessarily has a zero derivative.
Hence, we next seek to characterize this minimum in order to leverage it later in our proposed approach. 
For simplicity, we first only investigate a simplified version of Eq.(\ref{eqn:dts_objective}) that does not contain the problem-specific term $\Phi_{m}\left(z\right)$. This simplified version can be written as follows:
\begin{equation}
\begin{aligned}
& \underset{z}{\text{min}}~F\left(z\right) = 
& \underset{z}{\text{min}}
& & \dfrac{\lambda}{2} \sum_{i}  \underbrace{\left( z_{i} - \bar{z}_{N} \right)^{2}}_{= e\left( z_{i}, N\right)} + \sum_{i} c_{i} \left( z_{i} - t_{i}\right)^{2}.
\end{aligned}
\label{eqn:dts_objective_simplified}
\end{equation}
Taking the gradient of Eq.(\ref{eqn:dts_objective_simplified}) with respect to $z_i$ and setting it to zero provides
\begin{equation}
\nabla_{z_i} F  = 0 = \lambda \left( z_{i} - \bar{z}_{N} \right) + c_{i} \left( z_{i} - t_{i}\right).
\label{eqn:dts_gradient_definition}
\end{equation}
Hence, at the minima of Eq.(\ref{eqn:dts_objective_simplified}) we have
\begin{equation}
 z_{i} = \dfrac{\lambda \bar{z}_{N} + c_{i} t_{i}}{\lambda + c_{i}}.
\label{eqn:dts_gradient_solution}
\end{equation}
Now, we highlight the relation to the optimization function of the BL-Solver.
Its optimization objective is
\begin{equation}
\begin{aligned}
& \underset{z}{\text{min}}~F_{BL-Solver}\left(z\right) = 
& \underset{z}{\text{min}}
& & \dfrac{\lambda}{2} \sum_{i,j} W_{i,j} \left( z_{i} - {z}_{j} \right)^{2} + \sum_{i} c_{i} \left( z_{i} - t_{i}\right)^{2}.
\end{aligned}
\label{eqn:bs_solver_objective}
\end{equation}
Inspecting the derivative at the minimum as we did for Eq.(~\ref{eqn:dts_objective_simplified}) requires us to compute the gradient of Eq.(\ref{eqn:bs_solver_objective}) with respect to $z_i$ and setting it to zero:
\begin{equation}
\begin{aligned}
\nabla_{z_i} F_{BL-Solver}  = 0 &= 2 \lambda_{BL}\sum_{j} W_{i,j} \left( z_{i} - {z}_{j} \right) + c_{i} \left( z_{i} - t_{i}\right). \\
 z_{i} &= \dfrac{2\lambda_{BL} \sum_{j} W_{i,j}{z}_{j} + c_{i} t_{i}}{2\lambda_{BL} \sum_{j} W_{i,j} + c_{i}}.\\
 & = \dfrac{2\lambda_{BL}~\bar{z}_{N}  + \dfrac{c_{i} t_{i}}{\sum_{j} W_{i,j}}}{2\lambda_{BL} + \dfrac{c_{i}}{\sum_{j} W_{i,j}}}.
\end{aligned}
\label{eqn:bs_solver_gradient}
\end{equation}
The extra factor of 2 with $\lambda_{BL}$ is due to the fact that we have to consider the terms when the roles of $z_{i}$ and $z_{j}$ are exchanged.
The solutions in Eqn.(\ref{eqn:dts_gradient_solution}) and Eqn.(\ref{eqn:bs_solver_gradient}) look very similar.
The major difference is that in Eq.(\ref{eqn:bs_solver_gradient}), the contribution of confidence scores is weighted by $\sum_{j} W_{i,j}$ and hence it is edge-aware.
We also weigh the confidence $c_{i}$ during gradient descent updates by $\sum_{j} W_{i,j}$ to mimic its effect in Eq.(\ref{eqn:bs_solver_gradient}), which provides less weight to target $t_{i}$ when we have a large support via the similarities expressed by $W_{i,j}$.
We compute the confidence scores $c_{i}$ by estimating the variance of the $z$ in an edge-aware sense using the domain transform as suggested in~\cite{barron2016fast}:
\begin{equation}
c_{i}  = \exp{\left(-\dfrac{V_{i}}{2\sigma_{c}^{2}}\right)},~
V_{i} = DT\left(z_{i}^{2}\right) - DT\left(z_{i}\right)^{2}.
\label{eqn:confidence_scores}
\end{equation}
In this formulation, the domain transform is treated as a local estimate of the mean in an edge-aware sense, while $\sigma_{c}$ scales the variance to get confidence scores.

In summary, Eq.(\ref{eqn:dts_objective_simplified}) and Eq.(\ref{eqn:bs_solver_objective}) have the same optimal solution, but the solution of Eq.(\ref{eqn:dts_objective_simplified}) can be computed significantly faster by leveraging parallel computations.
The reason is that we replace the pair-wise term in Eq.(\ref{eqn:bs_solver_objective}) by the local edge-aware mean, which we can compute in an efficient manner (see Sec.~\ref{sec:domain_transform_derivation} for details), and we weigh the contribution of the target $t_{i}$s by adapting the input confidence according to the local support $\sum_{j} W_{i,j}$.
\subsection{Domain transform}
\label{sec:domain_transform_derivation}
Gastal and Oliveira~\cite{gastal2011domain} define an isometric transformation, which they call the domain transform (DT) for a 1-D multi-valued function $I : \Omega \rightarrow \mathbb{R}^{c},~ \Omega = [0,\infty)$ by treating $ C = \left(x,I(x)\right)$ as a curve in $\mathbb{R}^{c+1}$. The domain transform $DT : \mathbb{R}^{c+1} \rightarrow \mathbb{R}$ is such that it preserves distances between two points on the curve $C$ under a given norm.
Unlike Gastal and Oliveira~\cite{gastal2011domain}, we use the $L_2$ norm to define the distances, and hence we derive the domain transform here, which satisfies the constraint $\left\lVert DT(x_i,I(x_i)) - DT(x_j,I(x_j)) \right\rVert_2 = \left\lVert (x_i,I(x_i)) - (x_j,I(x_j)) \right\rVert_2$ for the nearest neighbors $x_i$ and $x_{i+1}$. This derivation follows closely Sec. 4 of Gastal and Oliveira~\cite{gastal2011domain}.
 Using a shorthand notation $ DT(x) = DT(x_i,I(x_i))$ and assuming a small shift $h$ in $x$, we can express the distance in pixels and color equal to the distance of the transform as follows:
\begin{equation}
\left(DT(x + h) - DT(x) \right)^{2} = h^{2} + \sum_{k=1}^{c}\left(I(x + h) - I(x) \right)^{2}.
\end{equation}
Taking the square root and constraining $DT(x)$ to be monotonic to avoid negative 
roots, followed by integrating both sides, we obtain
\begin{equation}
DT(u) = \int_{0}^{u} \sqrt{1 + \sum_{k=1}^{c} I_{k}'(x)}~dx, \quad u \in \Omega.
\end{equation}
%
%
Using this definition of the domain transform of the 4-D space $\left[X, R, G, B\right]$ with the curve C defined by RGB color and X denoting the domain, we can express the edge-aware mean as follows:
\begin{equation}
\bar{z}_{N_i} = \sum_{j\in N_i} H\left(DT(z_i),DT(z_j)\right) z_j.
\end{equation}
where $W_{i,j} = H\left(DT(z_i), DT(z_j)\right) = \delta\left\{ DT(z_i)-DT(z_j) \leq r\right\}$, and $\delta$ represents Dirac's delta function.
To see the relation with the simple domain transform blurring~\cite{gastal2011domain}, it can be seen that setting the confidence scores $c_{i}$ to zero in Eq.(\ref{eqn:dts_objective}) will lead to the same solution as the domain transform filtering.
Similarly, setting  $c_{i}$ to zero, and $W_{i.j}$ to Gaussian weights in color and space will lead to bilateral filtering.
Note that the above derivation is isometric since the function I is multi-valued but with a \textit{1-dimensional} domain $\Omega$.
By extending the domain to 2-D, the exact isometry is not valid, and Gastal and Oliveira~\cite{gastal2011domain} use alternating passes by separately considering the image as a function of X and then Y.

Now, all the terms except the application-specific terms in Eq.(\ref{eqn:dts_objective}) are defined.
In the following, we present how we apply our optimization framework to a variety of application scenarios where we adapt Eq.(\ref{eqn:dts_objective}) by changing function $\Phi_{m}$.
\vspace{-3mm}
\subsection{Stereo optimization}
\label{sec:stereo_optimization}
\begin{figure}
    \centering
    \begin{subfigure}[b]{0.2\textwidth}
        \centering
        \includegraphics[width=\textwidth] {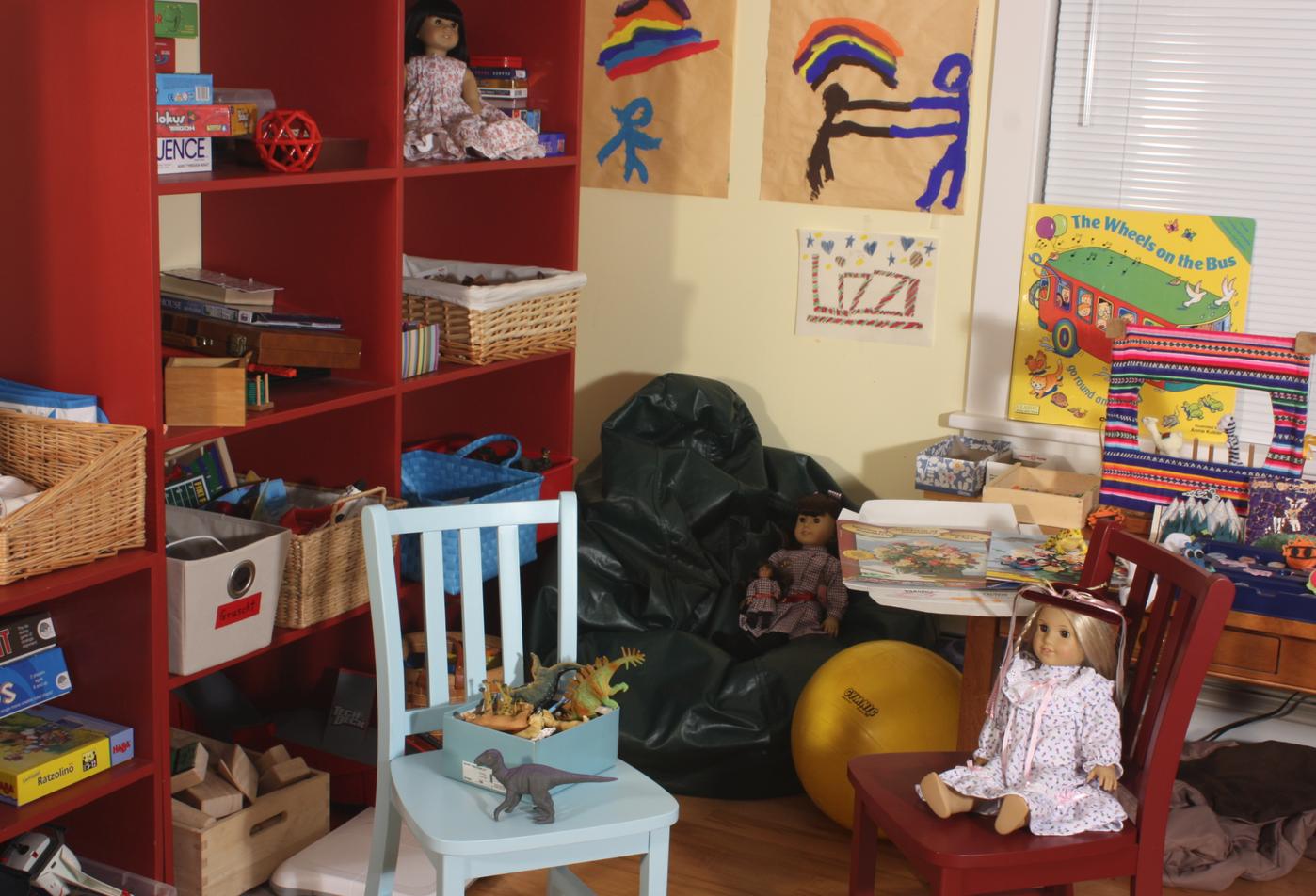}
        \caption{\centering Color image}
    \end{subfigure}%
    ~ \hspace{-1mm}
      \begin{subfigure}[b]{0.2\textwidth}
        \centering
        \includegraphics[width=\textwidth] {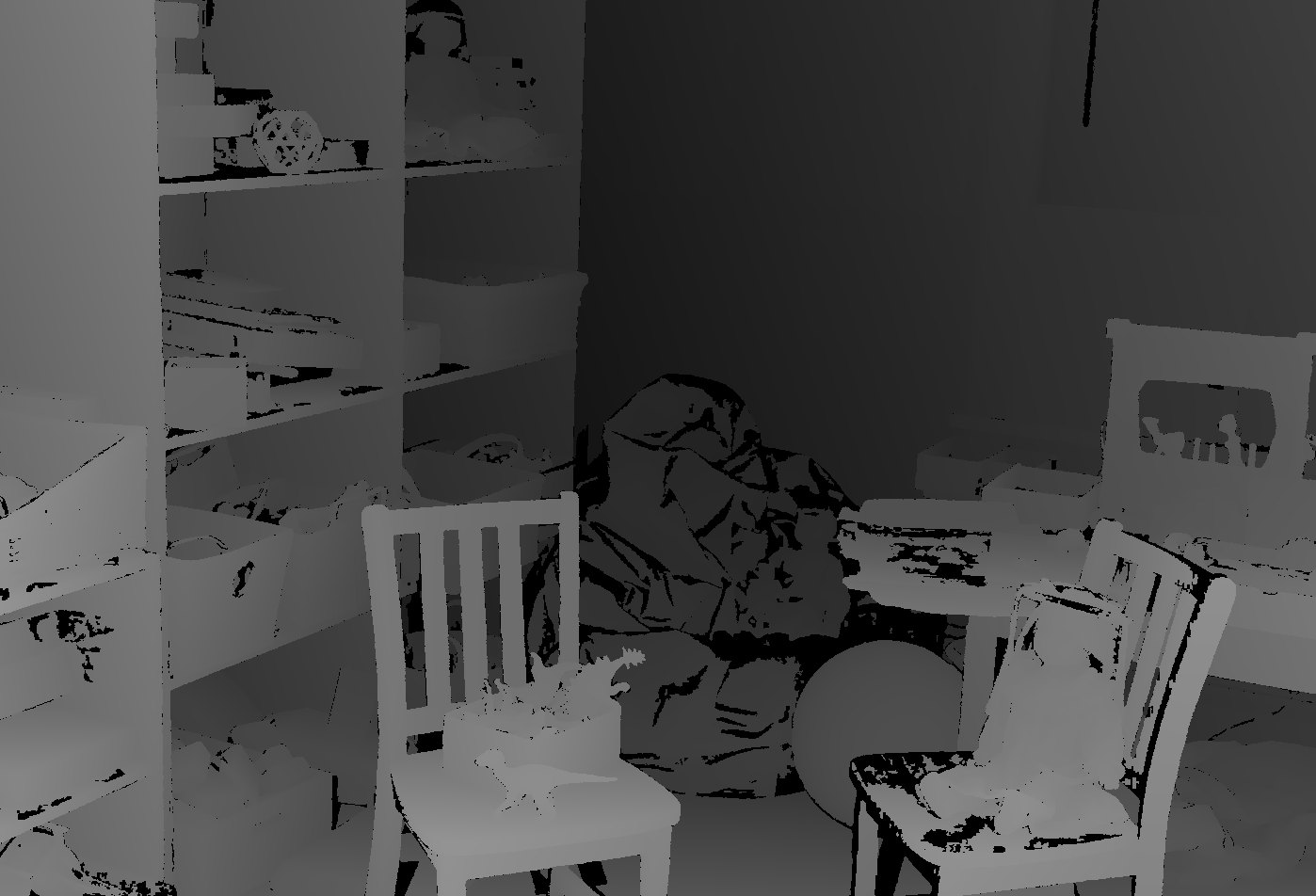}
        \caption{\centering GT disparity}
    \end{subfigure}%
    ~ \hspace{-1mm}
    \begin{subfigure}[b]{0.2\textwidth}
        \centering
        \includegraphics[width=\textwidth]{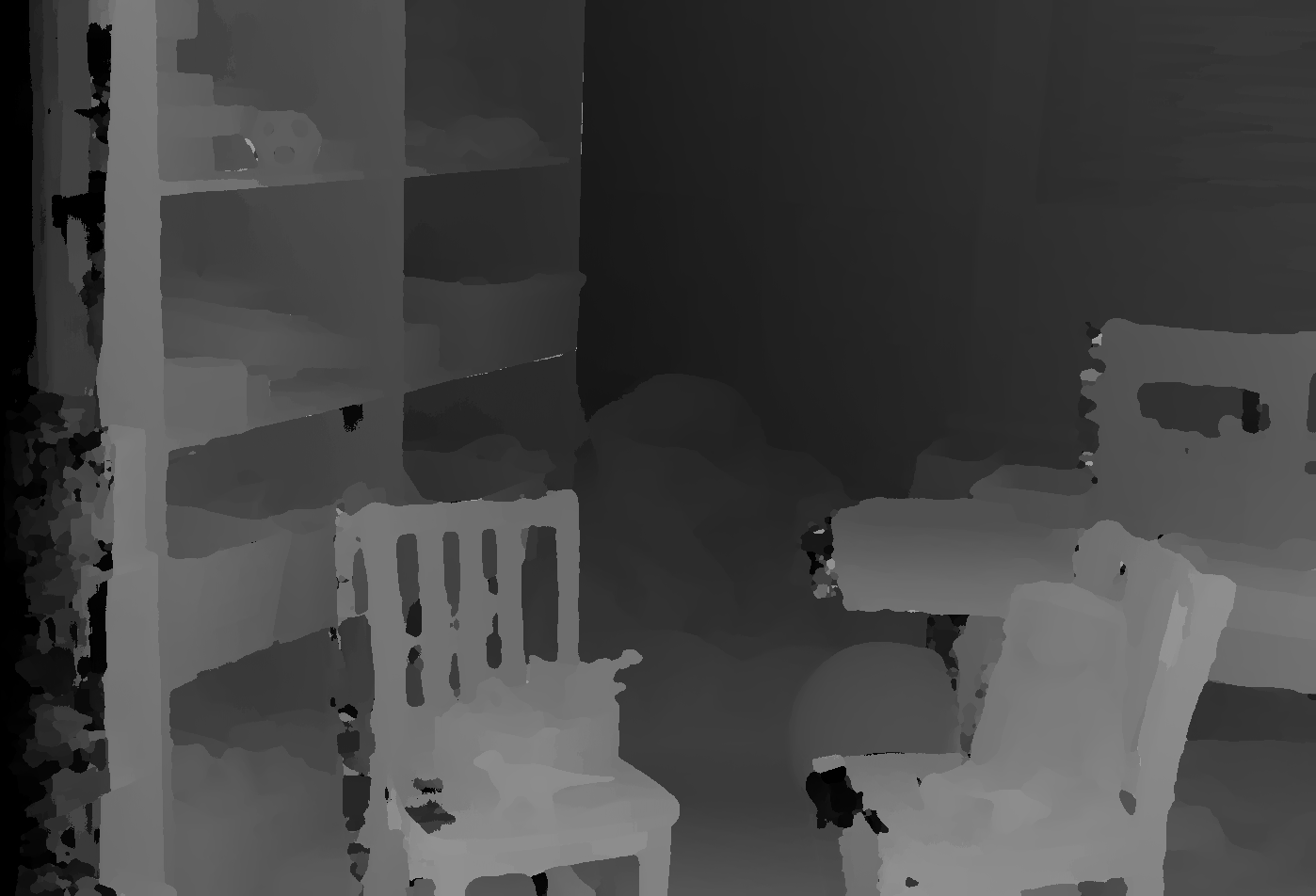}
        \caption{\centering Target image}
    \end{subfigure}%
    ~ \hspace{-1mm}
        \begin{subfigure}[b]{0.2\textwidth}
        \centering
        \includegraphics[width=\textwidth]{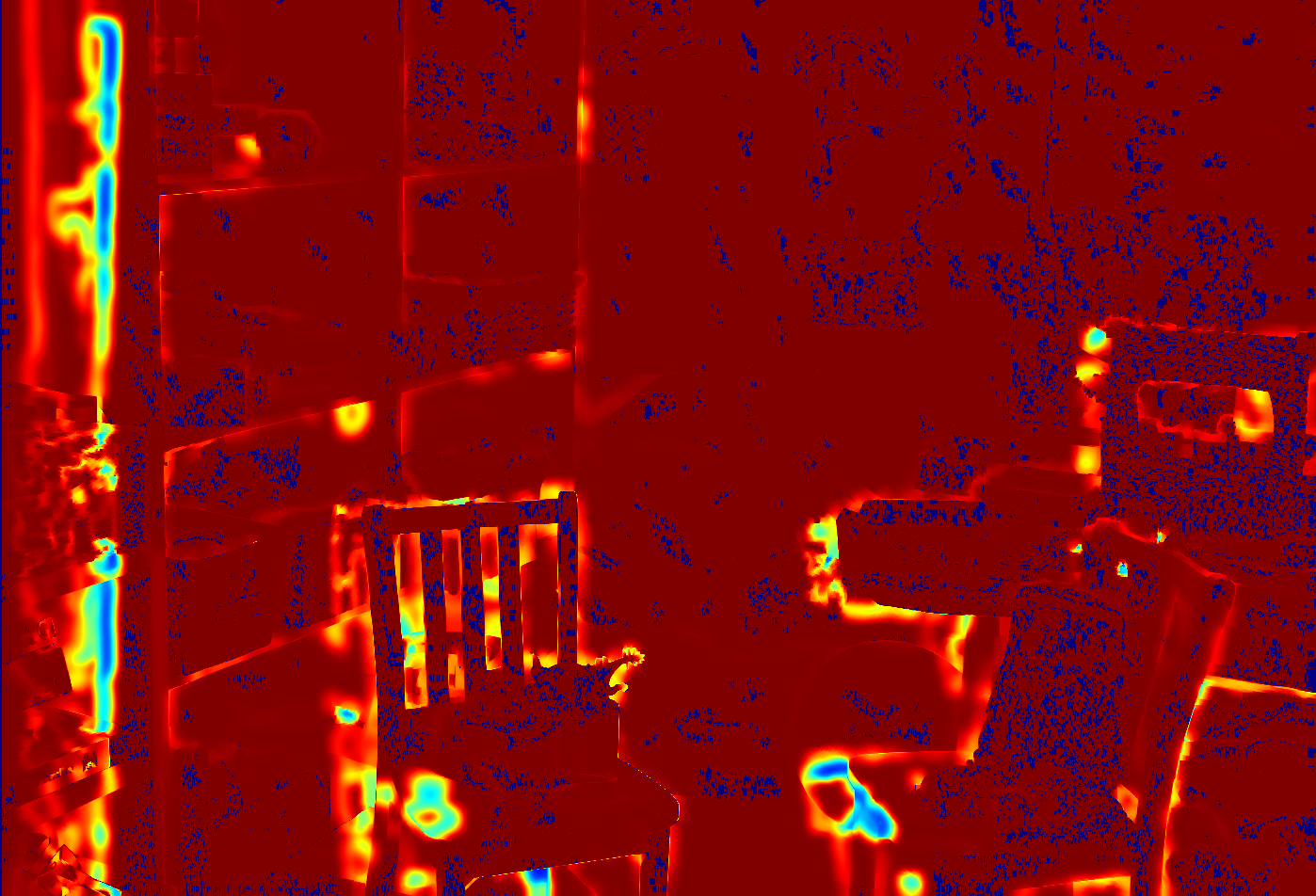}
        \caption{\centering Confidence}
    \end{subfigure}%
    ~ \hspace{-1mm}
        \begin{subfigure}[b]{0.2\textwidth}
        \centering
        \includegraphics[width=\textwidth]{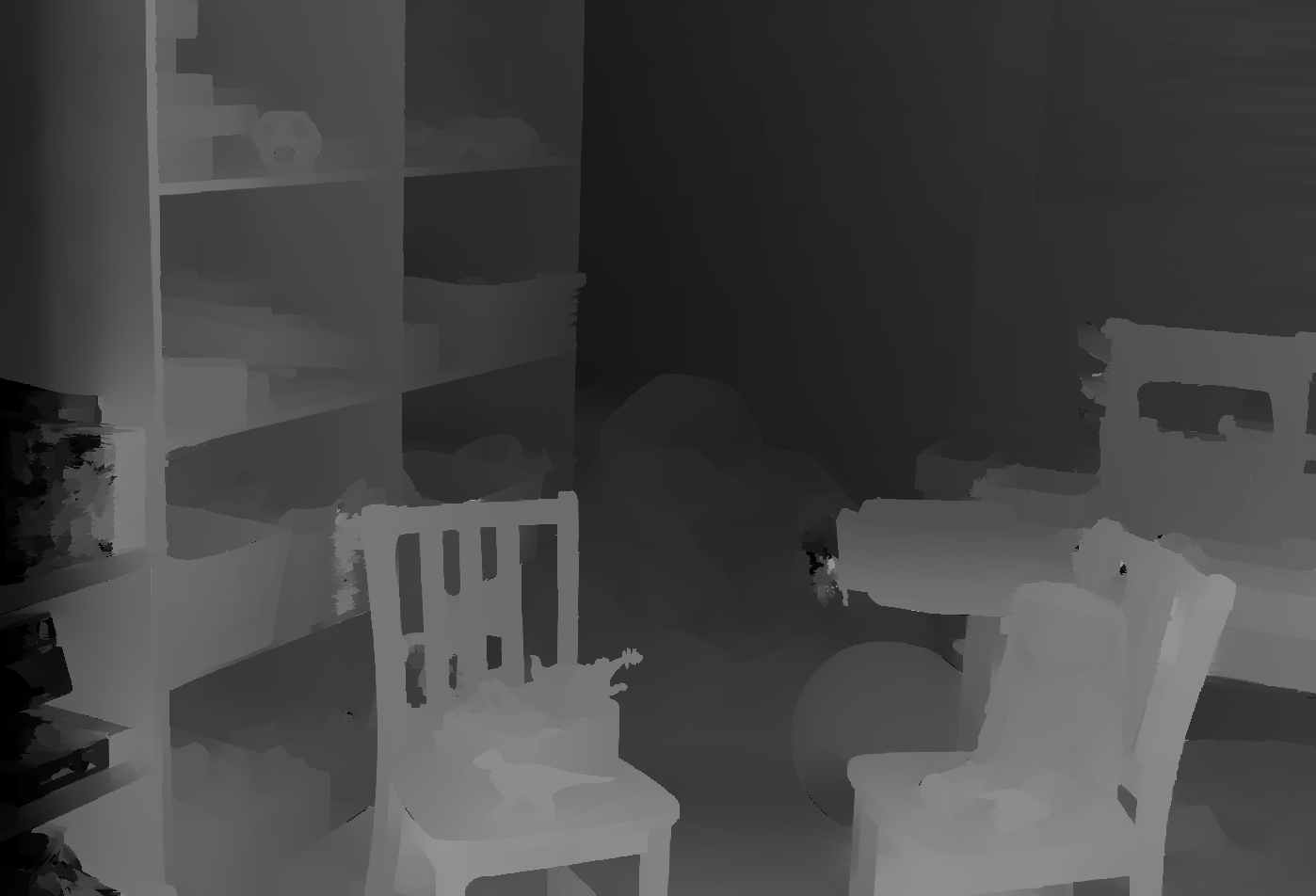}
        \caption{\centering Our result}
    \end{subfigure}%
    ~ \hspace{-1mm}

     \centering
    \begin{subfigure}[b]{0.2\textwidth}
        \centering
        \includegraphics[trim={10cm 5cm 25cm 17cm},clip,width=\textwidth] {images/Playroom_im0.jpg}
        \caption{\centering Color image}
    \end{subfigure}%
    ~ \hspace{-1mm}
          \begin{subfigure}[b]{0.2\textwidth}
        \centering
        \includegraphics[trim={10cm 5cm 25cm 17cm},clip,width=\textwidth] {images/Playroom_gt_disparity.png}
        \caption{\centering GT disparity}
    \end{subfigure}%
    ~ \hspace{-1mm}
    \begin{subfigure}[b]{0.2\textwidth}
        \centering
        \includegraphics[trim={10cm 5cm 25cm 17cm},clip,width=\textwidth]{images/Playroom_target.png}
        \caption{\centering Target image}
    \end{subfigure}%
    ~ \hspace{-1mm}
        \begin{subfigure}[b]{0.2\textwidth}
        \centering
        \includegraphics[trim={10cm 5cm 25cm 17cm},clip,width=\textwidth]{images/Playroom_optim_disparity.png}
        \caption{\centering Our result}
    \end{subfigure}%
    \caption{\label{fig:stereo_optimization} Stereo Optimization: The top row shows our result in (e) which is computed using the color image (a) used to define color distance in the domain transform, and target (c) disparity obtained from MC-CNN~\cite{zbontar2016stereo}. The confidence map (d) is used to weigh the target disparity in the optimization (Eq.(\ref{eqn:dts_objective})).  (f-i) show a zoomed area of (a-c,e). Notice in the zoomed regions that our results are aligned to the edges of the color image.}  
\end{figure}

\vspace{-1mm}
Stereo estimation is a well-studied problem~\cite{scharstein2002taxonomy,hirschmuller2007evaluation} in which the task is to estimate a matching correspondence of pixels in the left image to the pixels in the right image.
This matching correspondence defines the disparity of the pixels and in turn the depth, and when done for each pixel provides us with a disparity map.
Typically, dense search is done along the row of a rectified pair by matching the pixel color similarity known as photometric matching cost.
In the following, we refine a disparity map.
We obtain the disparity map from MC-CNN~\cite{zbontar2016stereo}, which acts as our target (Fig.~\ref{fig:stereo_optimization}(c)), for which we calculate a confidence score (shown in Fig.~\ref{fig:stereo_optimization}(d)) using Eq.(\ref{eqn:confidence_scores}).
We use the left color image to define and compute the edge-aware mean and optimize the disparities to obtain a disparity map that is smooth at homogeneous regions but has crisp edges (Fig.~\ref{fig:stereo_optimization}(e)).
Similar to our proposed solver, Barron and Poole~\cite{barron2016fast} show that the BL-Solver works well for a wide variety of optimization problems including stereo.
When they apply the BL-Solver to the stereo problem, they achieve faster convergence compared to BL-Stereo because they neglect the physical implication of changing the disparity.
In other words, if an optimizer changes the estimate of disparity at a point in left image, this gets reflected in a change in the color of its matching pixel in the right image.
Here, we present a method for solving for the disparity in an edge-aware sense while having a photometric penalty for the left-right matching.
Our loss for stereo optimization is as follows:
\begin{equation}
\begin{aligned}
& \underset{z}{\text{min}}
& & \dfrac{\lambda}{2} \sum_{i}  \left( z_{i} - \bar{z}_{N} \right)^{2} + \sum_{i} c_{i} \underbrace{\left(z_{i} - t_{i}\right)^{2}}_{target~term} + \dfrac{1}{2} \sum_{i} \underbrace{\gamma \left( I_{L}(i) - I_{R}(i - z_i) \right)^{2}}_{ = \lambda_{m}\Phi_{m}\left(z_{i}\right)},
\end{aligned}
\label{eqn:domain_transform_optim_stereo}
\end{equation}
where $I_{L}$ is the left image and $I_{R}$ is the right image of the stereo pair.
For robust optimization, we use a Charbonnier loss $\rho\left(r\right) = \sqrt{r^{2} + \varepsilon^{2}}$ with $\varepsilon = 0.001$ on the target term $r = \left( z_{i} - t_{i}\right)$, which has been shown to be effective for optical flow~\cite{sun2010secrets}.
We use Zbontar and LeCun's MC-CNN~\cite{zbontar2016stereo} as the target for our stereo optimization.
Next, we will detail the application of our method to the problem of rendering defocus from depth, which is  another application heavily relying on accurate depth edges.
\subsection{Synthetic defocus from depth}
\label{sec:synthetic_defocus}
\begin{figure}
    \centering
         \begin{subfigure}[b]{0.33\textwidth}
        \centering
        \includegraphics[width=\textwidth] {images/Jadeplant_im0.jpg}
        \caption{Original Image}
    \end{subfigure}%
    ~
    \begin{subfigure}[b]{0.33\textwidth}
        \centering
        \includegraphics[width=\textwidth] {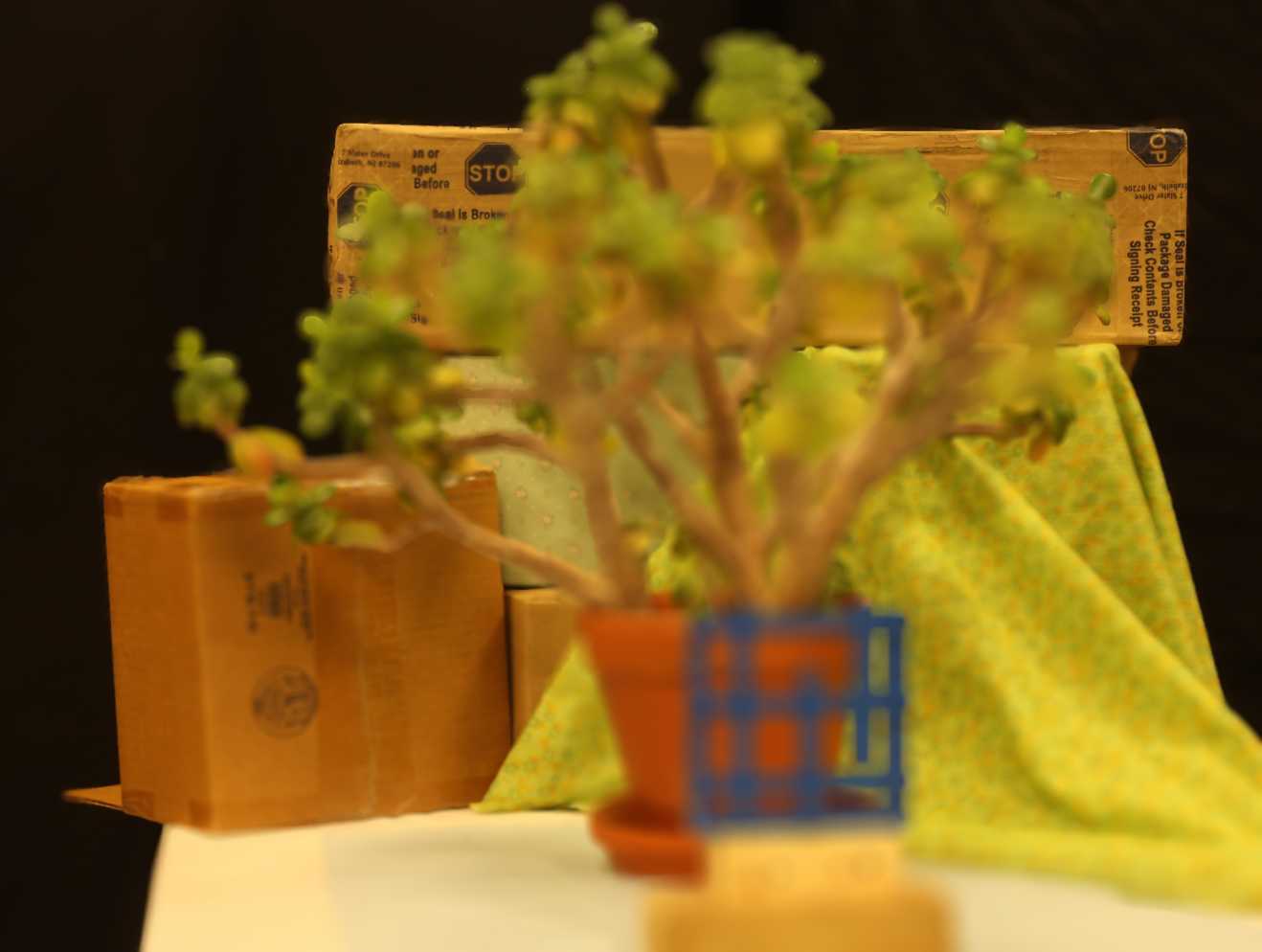}
        \caption{Ours}
    \end{subfigure}%
    ~ 
      \begin{subfigure}[b]{0.33\textwidth}
        \centering
        \includegraphics[width=\textwidth] {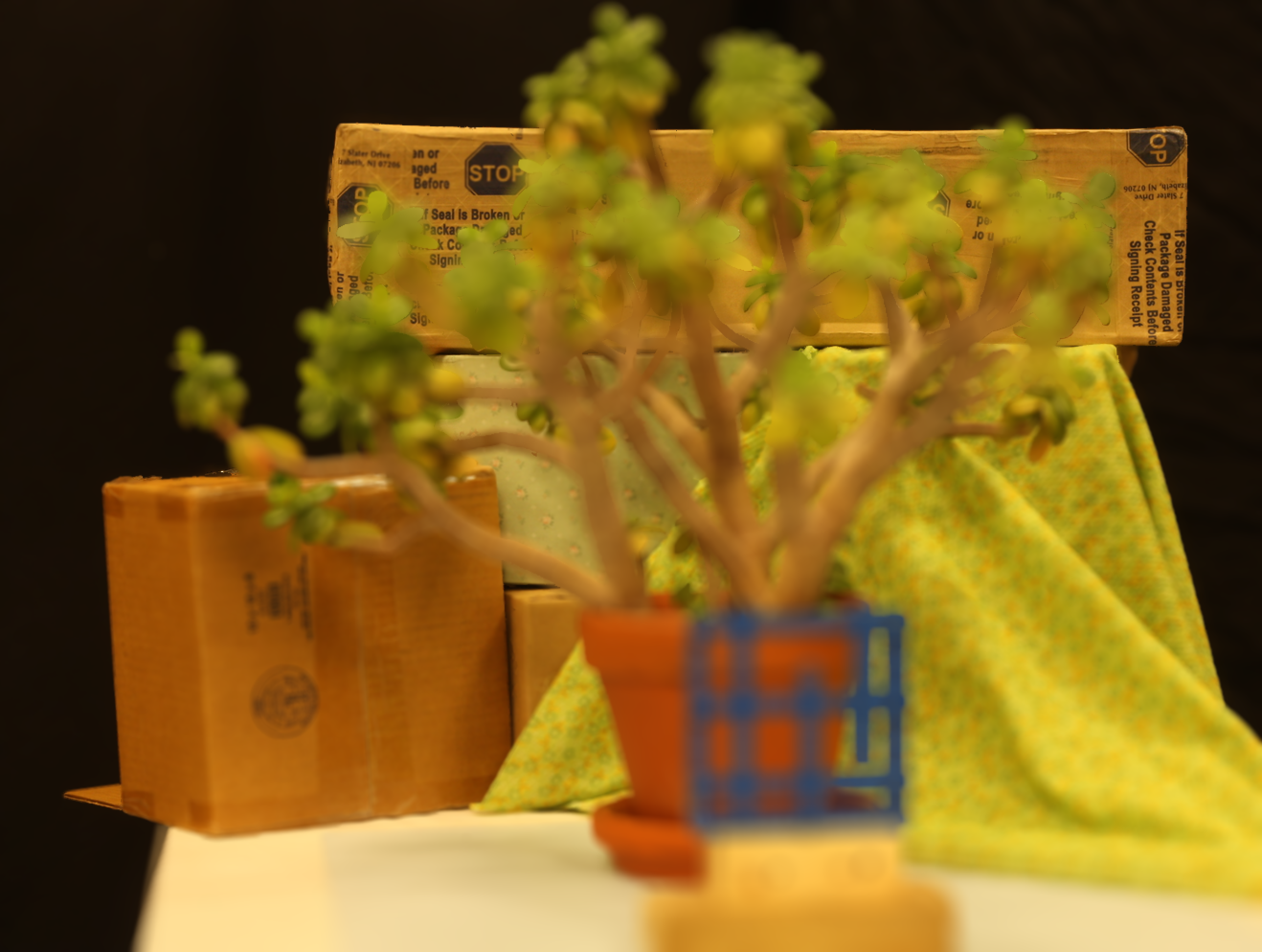}
        \caption{GT disparity}
    \end{subfigure}%
    \caption{\label{fig:synthetic_defocus_jadeplant_far}Render from defocus for the Middlebury Jadeplant scene. (a) Original color image. (b) Our result where the background is in focus. (c) Result computed using the ground-truth disparity.}
    
    \setlength{\fboxrule}{0.2pt}
       \centering
    \begin{subfigure}[b]{0.33\textwidth}
       \centering
    \stackinset{l}{}{b}{}
  {\fcolorbox{red}{red}{\includegraphics[trim={15cm 20cm 20cm 4cm},clip,width=0.6in]{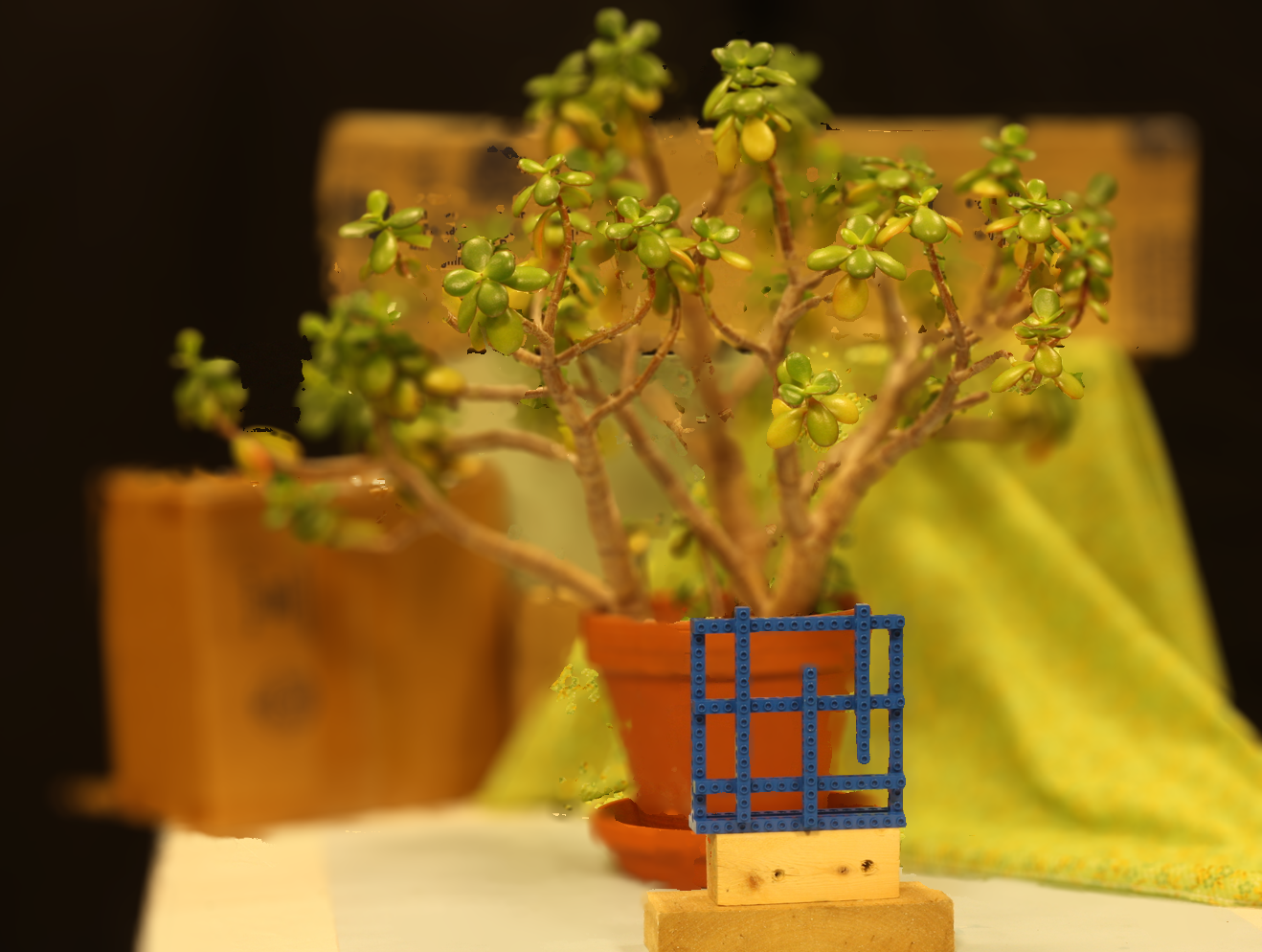}}}
  {\includegraphics[width=\textwidth]{images/Jadeplant_defocus_target.png}}
\caption{MC-CNN}
    \end{subfigure}%
    ~
     \begin{subfigure}[b]{0.33\textwidth}
       \centering
     \stackinset{l}{}{b}{}
  {\fcolorbox{red}{red}{\includegraphics[trim={15cm 20cm 20cm 4cm},clip,width=0.6in]{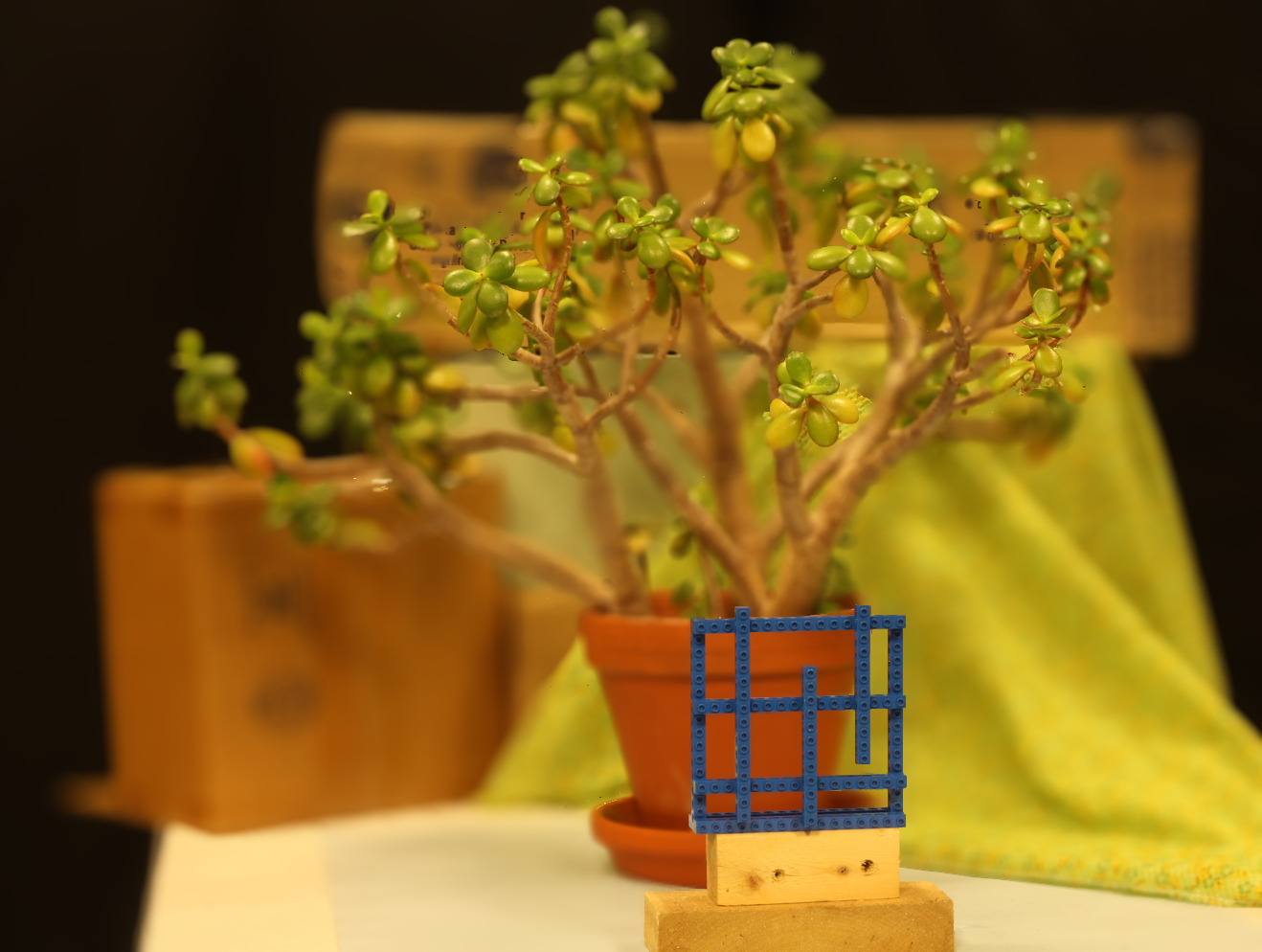}}}
  {\includegraphics[width=\textwidth]{images/Jadeplant_defocus_optim.png}}
\caption{Ours}
    \end{subfigure}%
        ~ 
         \begin{subfigure}[b]{0.33\textwidth}
           \centering
         \stackinset{l}{}{b}{}
  {\fcolorbox{red}{red}{\includegraphics[trim={15cm 20cm 20cm 4cm},clip,width=0.6in]{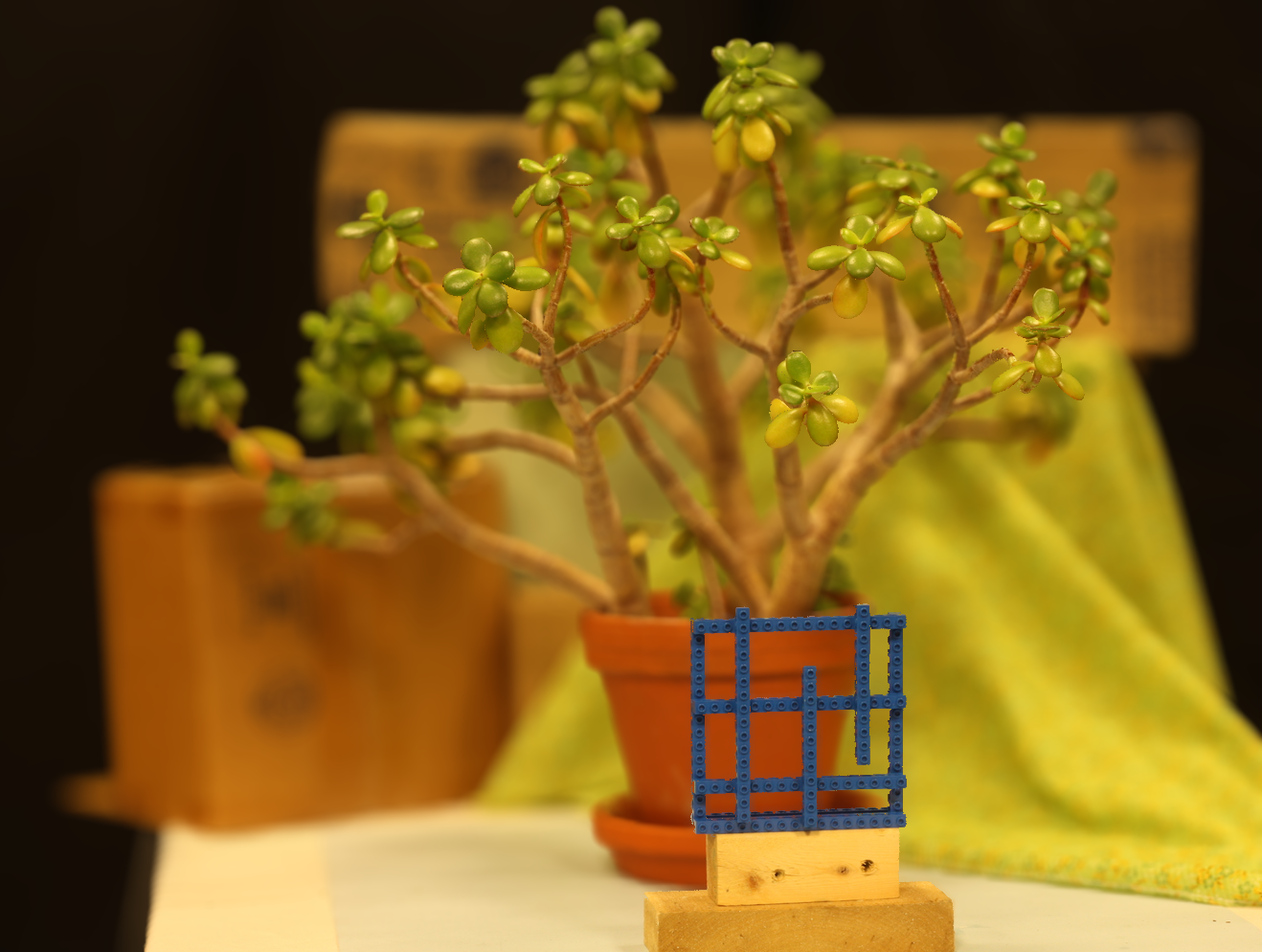}}}
  {\includegraphics[width=\textwidth]{images/Jadeplant_defocus_gt.png}}
\caption{GT disparity}
    \end{subfigure}%
    \caption{\label{fig:synthetic_defocus_jadeplant_near}Render from defocus for the Middlebury Jadeplant scene. (a) Results obtained using the MC-CNN depthmap where the front blue block is in focus. (b) Our result. (c) Result computed using the ground-truth disparity. The inset shows details and highlights improvements around the edges.}
    
     \centering
 \begin{subfigure}[b]{0.33\textwidth}
        \centering
        \includegraphics[width=\textwidth] {images/Playroom_im0.jpg}
        \caption{Original Image}
    \end{subfigure}%
        ~ 
    \begin{subfigure}[b]{0.33\textwidth}
        \centering
        \includegraphics[width=\textwidth] {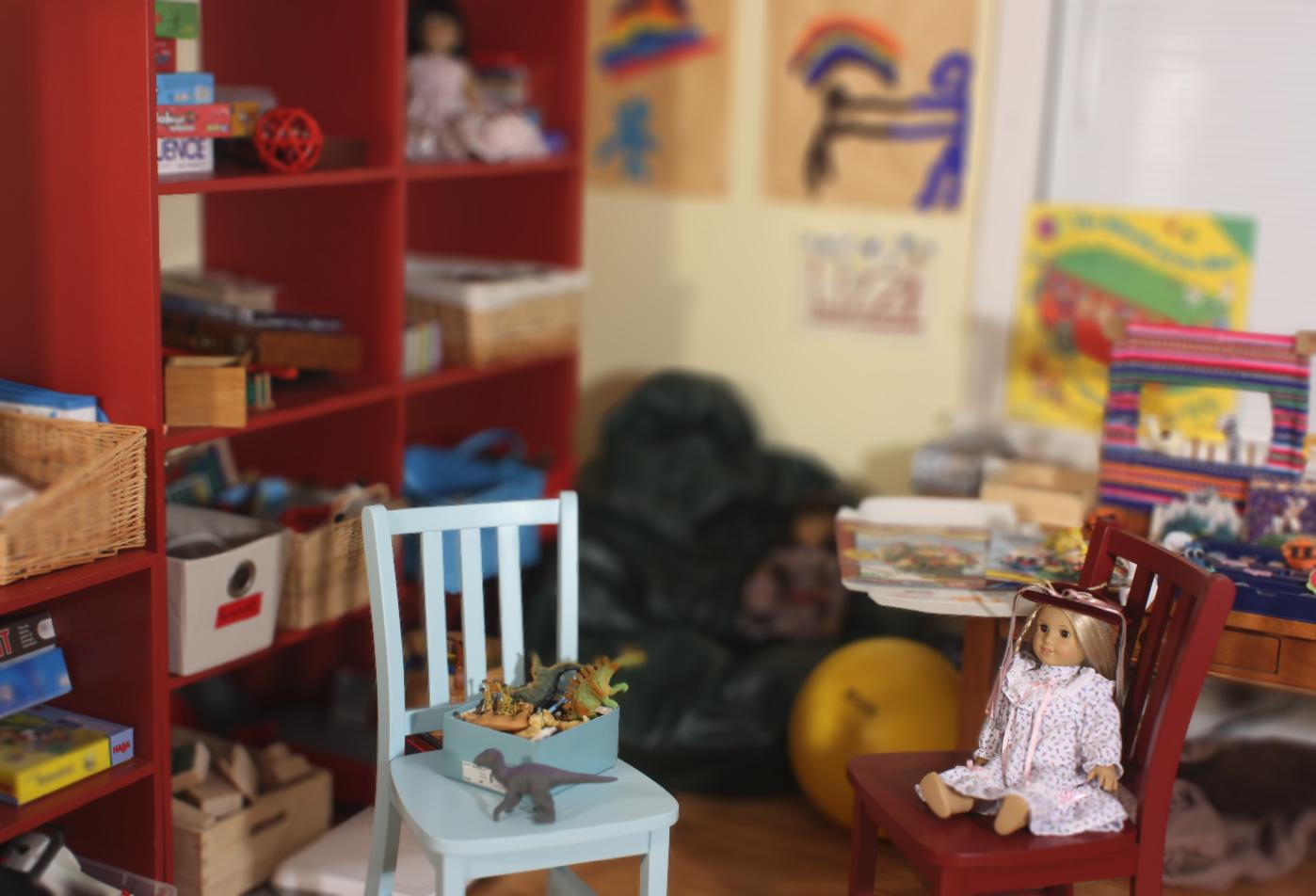}
        \caption{Ours}
    \end{subfigure}%
    ~ 
      \begin{subfigure}[b]{0.33\textwidth}
        \centering
        \includegraphics[width=\textwidth] {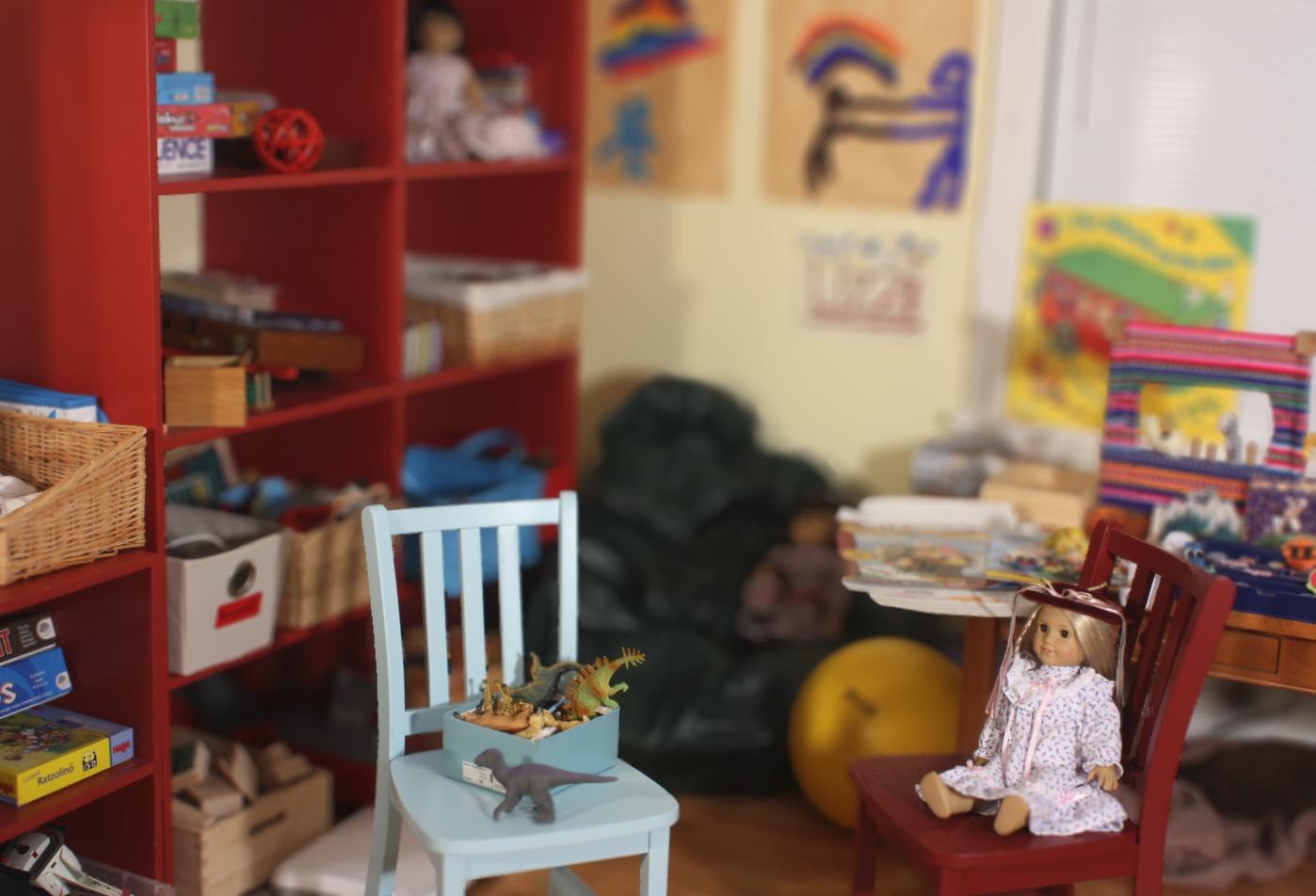}
        \caption{GT disparity}
    \end{subfigure}%
    \caption{\label{fig:synthetic_defocus_playroom}Render from defocus for the Middlebury Playroom scene. (a) Original color image. (b) Our result where the chairs in the front are in focus. (c) Result computed using the ground-truth disparity.}
\end{figure}

Interest in creating synthetic defocus from depth is growing, with phones like the Google Pixel 2 and the OnePlus 5T providing a portrait mode where the shallow depth of field effect is mimicked through the estimation of depth.
BL-Stereo's synthetic defocus method is used as part of the Lens Blur feature on Google's phones~\cite{barron2015fast}.
We use our stereo optimization from Sec.~\ref{sec:stereo_optimization}  to estimate depth maps, which retain sharp discontinuities at color edges.
Figs.~\ref{fig:synthetic_defocus_jadeplant_far},~\ref{fig:synthetic_defocus_jadeplant_near}, and~\ref{fig:synthetic_defocus_playroom} show the original color image and the defocus rendering produced by using our estimated depthmaps and the ground-truth depthmaps for scenes in the Middlebury dataset~\cite{scharstein2014high}.
As our stereo optimization is edge-aware, the defocus rendering maintains high quality even at the edges.
Notice that in the insets of Fig.~\ref{fig:synthetic_defocus_jadeplant_near}., MC-CNN has jarring artifacts, especially at the edges, while the rendering using our estimated depthmap is more smooth.
In the Jadeplant scene shown in Fig.~\ref{fig:synthetic_defocus_jadeplant_far}, the background is in focus, and for the same scene Fig.~\ref{fig:synthetic_defocus_jadeplant_near} the blue block in the front is kept in focus. 
In the Playroom scene illustrated in Fig.~\ref{fig:synthetic_defocus_playroom}, the front chairs are chosen to be in focus.
To render the synthetic defocus, we used the algorithm described in Sec. 6 of the supplementary material of Barron~\etal~\cite{barron2015fast}.
This shows that our results are qualitatively better than MC-CNN, and the most noticeable improvements are because we optimize in an edge-aware sense.
\subsection{Depth super-resolution}
\label{sec:depth_super_resolution}
\begin{figure}
    \captionsetup[subfigure]{font=scriptsize,labelfont=scriptsize}

 \setlength{\fboxrule}{0.2pt}
       \centering
    \begin{subfigure}[t]{0.25\textwidth}
       \centering
    \stackinset{l}{}{b}{}
  {\fcolorbox{red}{red}{\includegraphics[trim={17cm 22cm 20cm 6cm},clip,width=0.6in]{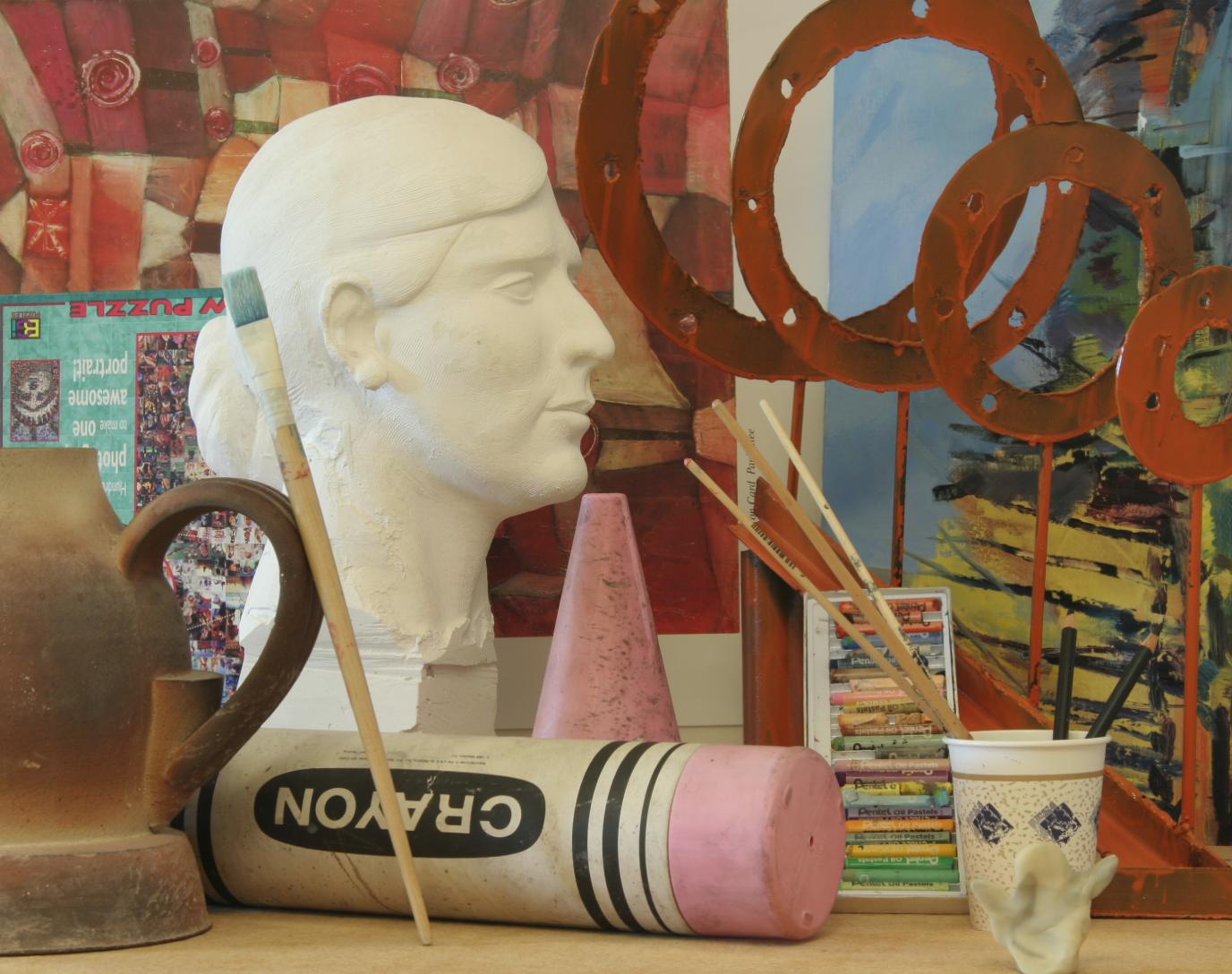}}}
  {\includegraphics[width=\textwidth]{images/art_big_color_loaded.jpg}}
\caption{\centering Color image.}
    \end{subfigure}%
    ~\hspace{-1mm}
     \begin{subfigure}[t]{0.25\textwidth}
       \centering
     \stackinset{l}{}{b}{}
  {\fcolorbox{red}{red}{\includegraphics[trim={17cm 22cm 20cm 6cm},clip,width=0.6in]{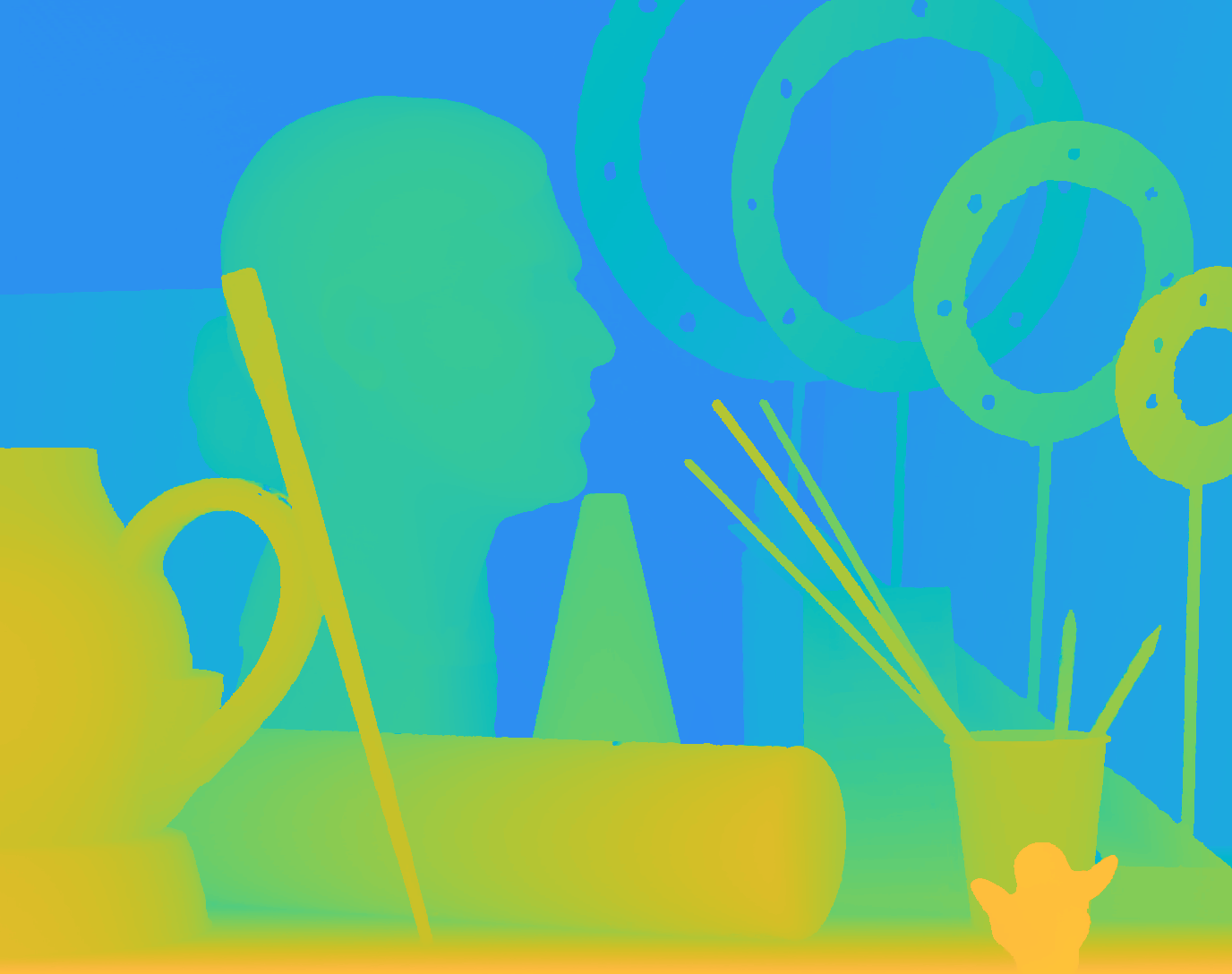}}}
  {\includegraphics[width=\textwidth]{images/Art_depth_gt_map.png}}
\caption{\centering GT disparity}
    \end{subfigure}%
~\hspace{-1mm}
\begin{subfigure}[t]{0.25\textwidth}
        \centering
         \stackinset{l}{}{b}{}
  {\fcolorbox{red}{red}{\includegraphics[trim={17cm 22cm 20cm 6cm},clip,width=0.6in]{images/output_4_n_map.png}}}
  {\includegraphics[width=\textwidth]{images/output_4_n_map.png}}
\caption{\centering Our result}
    \end{subfigure}%
    ~\hspace{-1mm}
    \begin{subfigure}[t]{0.25\textwidth}
        \centering
         \stackinset{l}{}{b}{}
  {\fcolorbox{red}{red}{\includegraphics[trim={17cm 22cm 20cm 6cm},clip,width=0.6in]{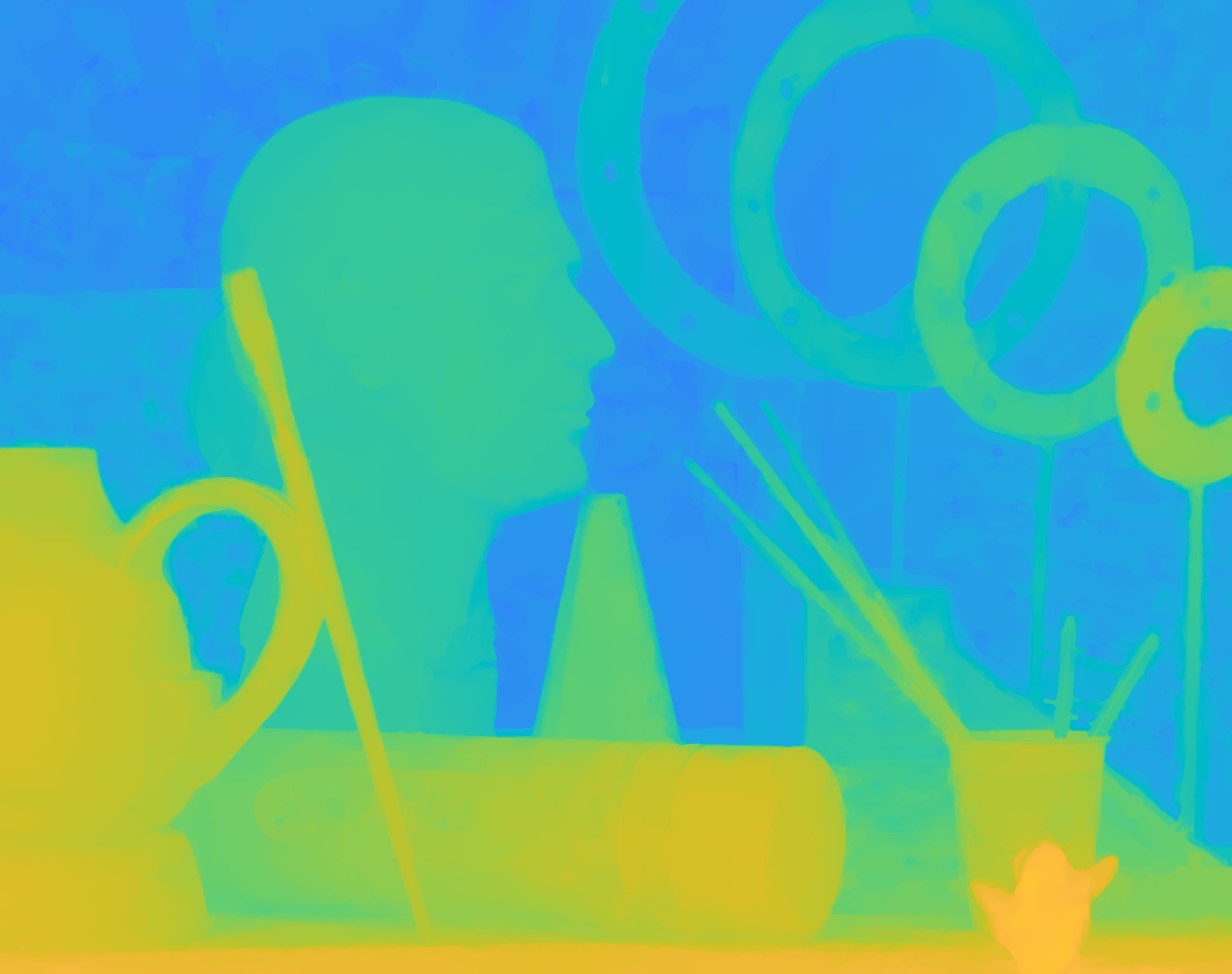}}}
  {\includegraphics[width=\textwidth]{images/output_4_n_barronpoole_map.png}}
\caption{\centering Barron and Poole~\cite{barron2016fast}.}
    \end{subfigure}%
    
    \caption{\label{fig:depth_super_resolution}Depth super-resolution: (a) shows original color image, (b) ground truth disparity, (c) our optimized disparity and (d) results using BL-Solver obtained from the author's website~\cite{jon_barron_website}. The inset highlights the details and the amount of smoothness we obtain in homogeneous regions while being edge-aware.}
\end{figure}

The availability of cheap commodity depth sensors like the Microsoft Kinect, Asus Xtion, and Intel RealSense has spurred many avenues of research, including depth super-resolution.  
Depth super-resolution is important for sensors like these because, often, the color camera is of high resolution, but the  depth camera/projector has low-resolution, which leads to crude depth maps~\cite{khoshelham2012accuracy}.
Ferstl~\etal~\cite{ferstl2013image} adapted the Middlebury dataset for the depth super-resolution task to create a benchmark, on which we evaluate our method, here.
For this task, we use simple bicubic interpolation for upsampling the low-resolution depth map and use this map as a target in our optimization; we use the high-resolution color image to compute the domain transform based edge-aware mean and obtain our optimized result (Fig.~\ref{fig:depth_super_resolution}(d)).
We follow Barron and Poole~\cite{barron2016fast} by setting the confidence scores using a Gaussian bump model to represent the contribution of each pixel to the nearby upsampled pixels.
We do not use additional penalties in Eq.(\ref{eqn:dts_objective}) for this task in the form of $\Phi_{m}$.

%% file: results.tex
\vspace{-3mm}
\section{Experiments}
\label{sec:results}
\begin{figure}
    \centering
    \captionsetup[subfigure]{font=scriptsize,labelfont=scriptsize}
    \begin{subfigure}[t]{0.17\textwidth}
        \centering
        \includegraphics[trim={25cm 0cm 10cm 20cm},clip,width=\textwidth] {images/Jadeplant_im0.jpg}
        \caption{\centering Color image}
    \end{subfigure}%
    ~ 
          \begin{subfigure}[t]{0.17\textwidth}
        \centering
        \includegraphics[trim={25cm 0cm 10cm 20cm},clip,width=\textwidth] {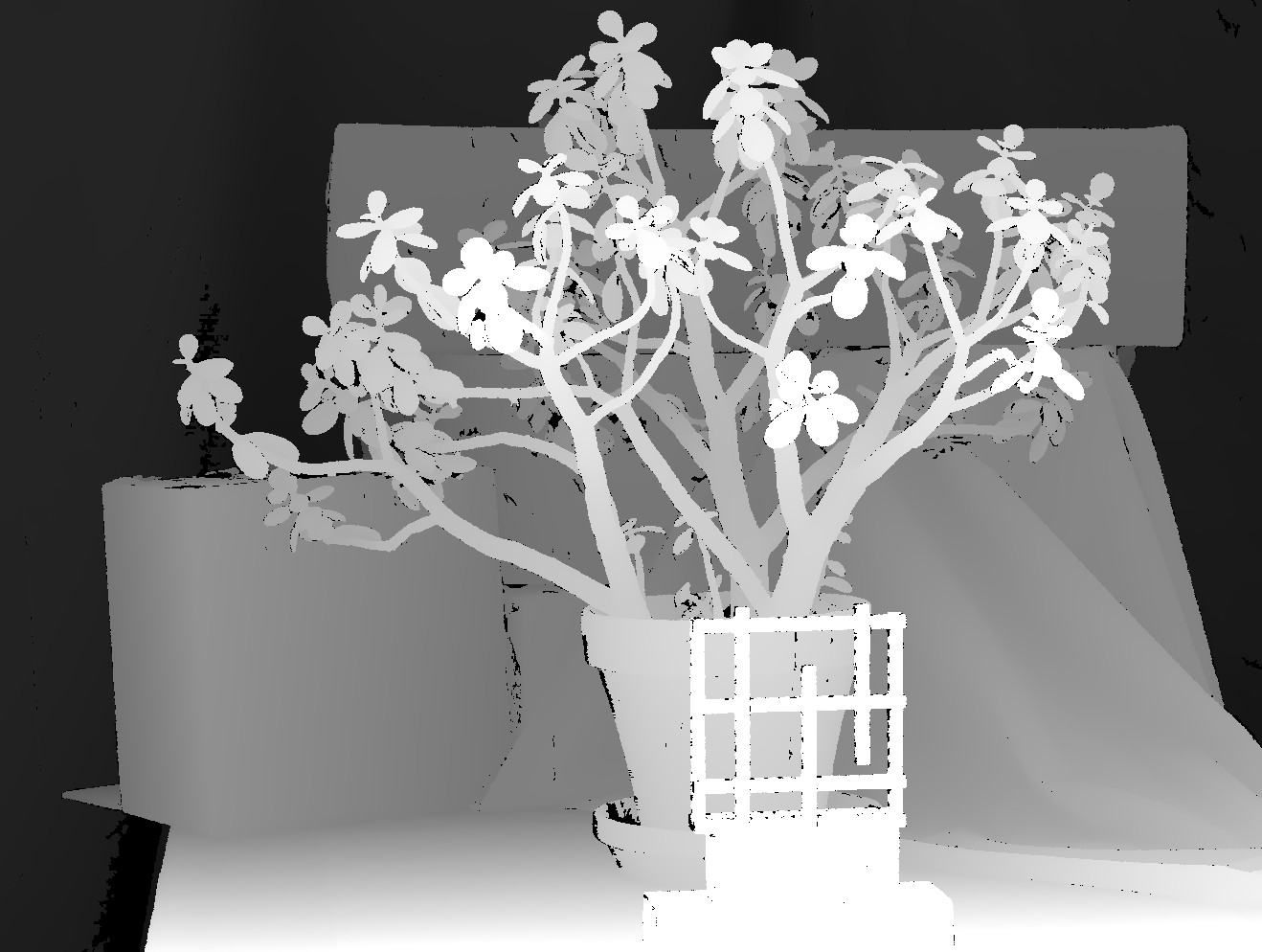}
        \caption{\centering GT disparity}
    \end{subfigure}%
    ~ 
    \begin{subfigure}[t]{0.17\textwidth}
        \centering
        \includegraphics[trim={25cm 0cm 10cm 20cm},clip,width=\textwidth]{images/Jadeplant_target.png}
        \caption{\centering Target image}
    \end{subfigure}%
    ~ 
        \begin{subfigure}[t]{0.17\textwidth}
        \centering
        \includegraphics[trim={25cm 0cm 10cm 20cm},clip,width=\textwidth]{images/Jadeplant_optim_disparity.png}
        \caption{\centering Our result}
    \end{subfigure}%
    \caption{\label{fig:stereo_jadeplant_closeup}Example stereo optimization results for a closeup of the Middlebury Jadeplant scene. (a) Color image. (b) Ground-truth disparity. (c) Target obtained using MC-CNN~\cite{zbontar2016stereo}. (d) Our result. }

     \centering
    \begin{subfigure}[b]{0.2\textwidth}
        \centering
        \includegraphics[trim={25cm 0cm 10cm 20cm},clip,width=\textwidth] {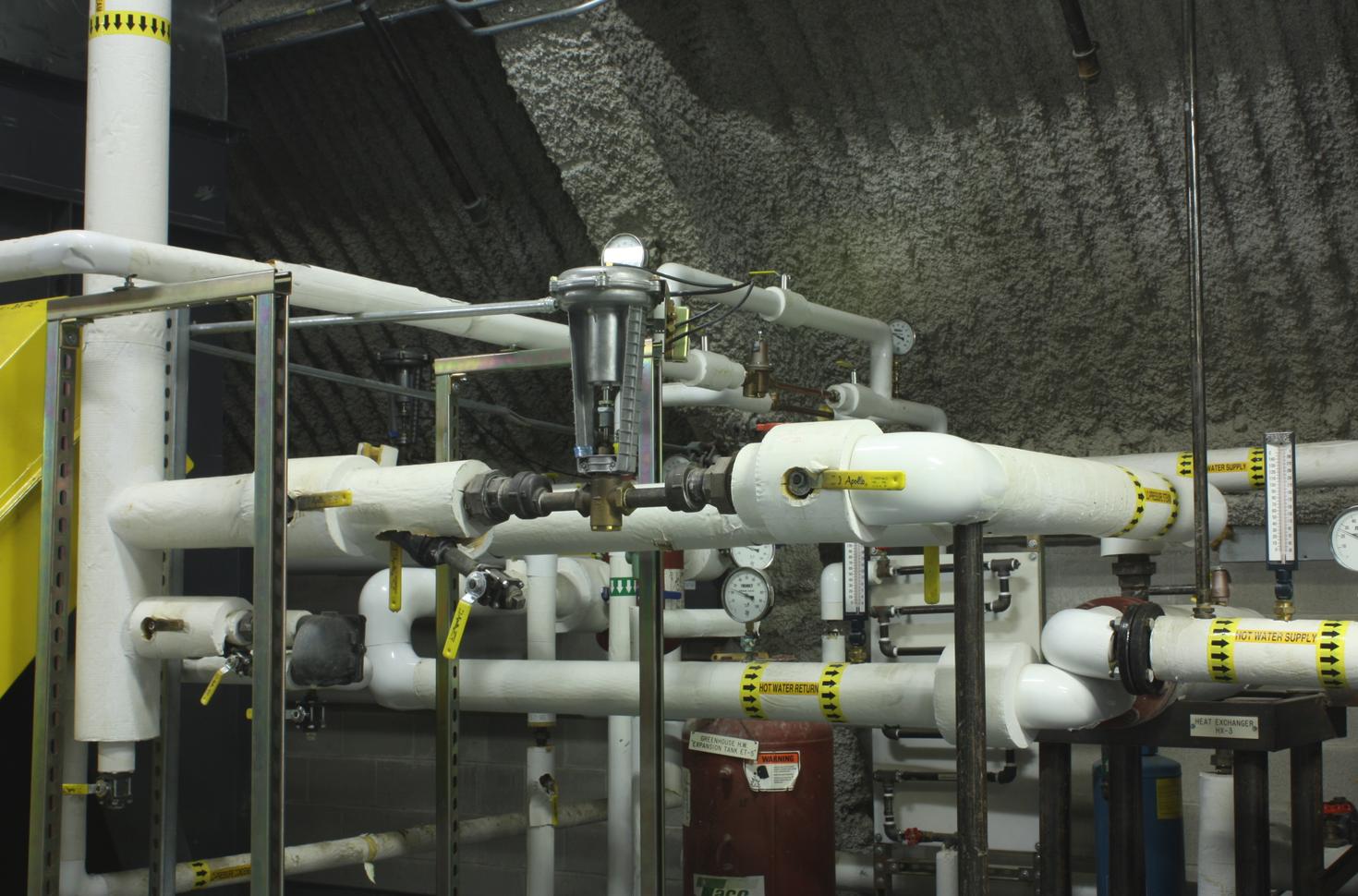}
        \caption{\centering Color image}
    \end{subfigure}%
    ~ 
          \begin{subfigure}[b]{0.2\textwidth}
        \centering
        \includegraphics[trim={25cm 0cm 10cm 20cm},clip,width=\textwidth] {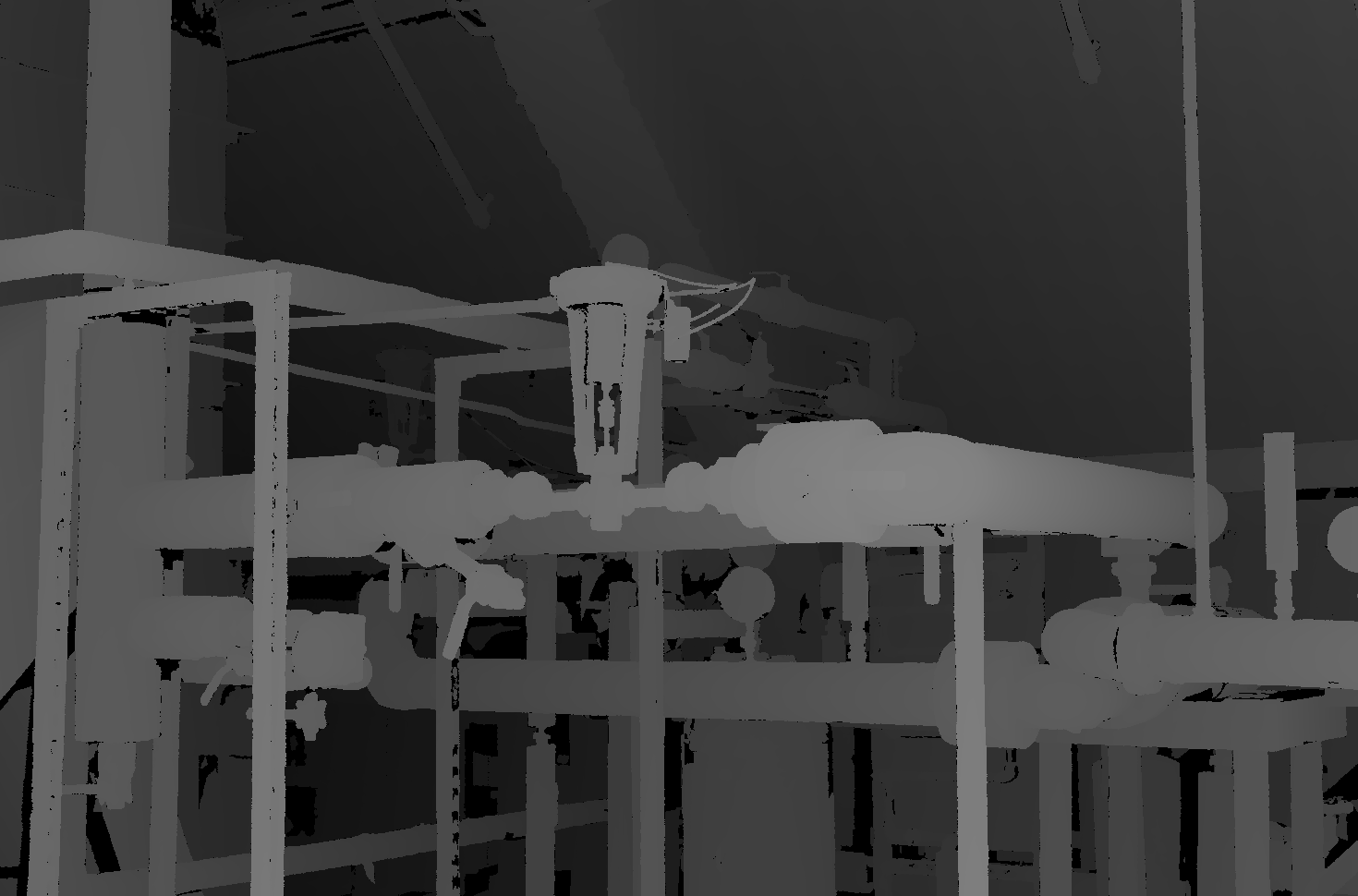}
        \caption{\centering GT disparity}
    \end{subfigure}%
    ~ 
    \begin{subfigure}[b]{0.2\textwidth}
        \centering
        \includegraphics[trim={25cm 0cm 10cm 20cm},clip,width=\textwidth]{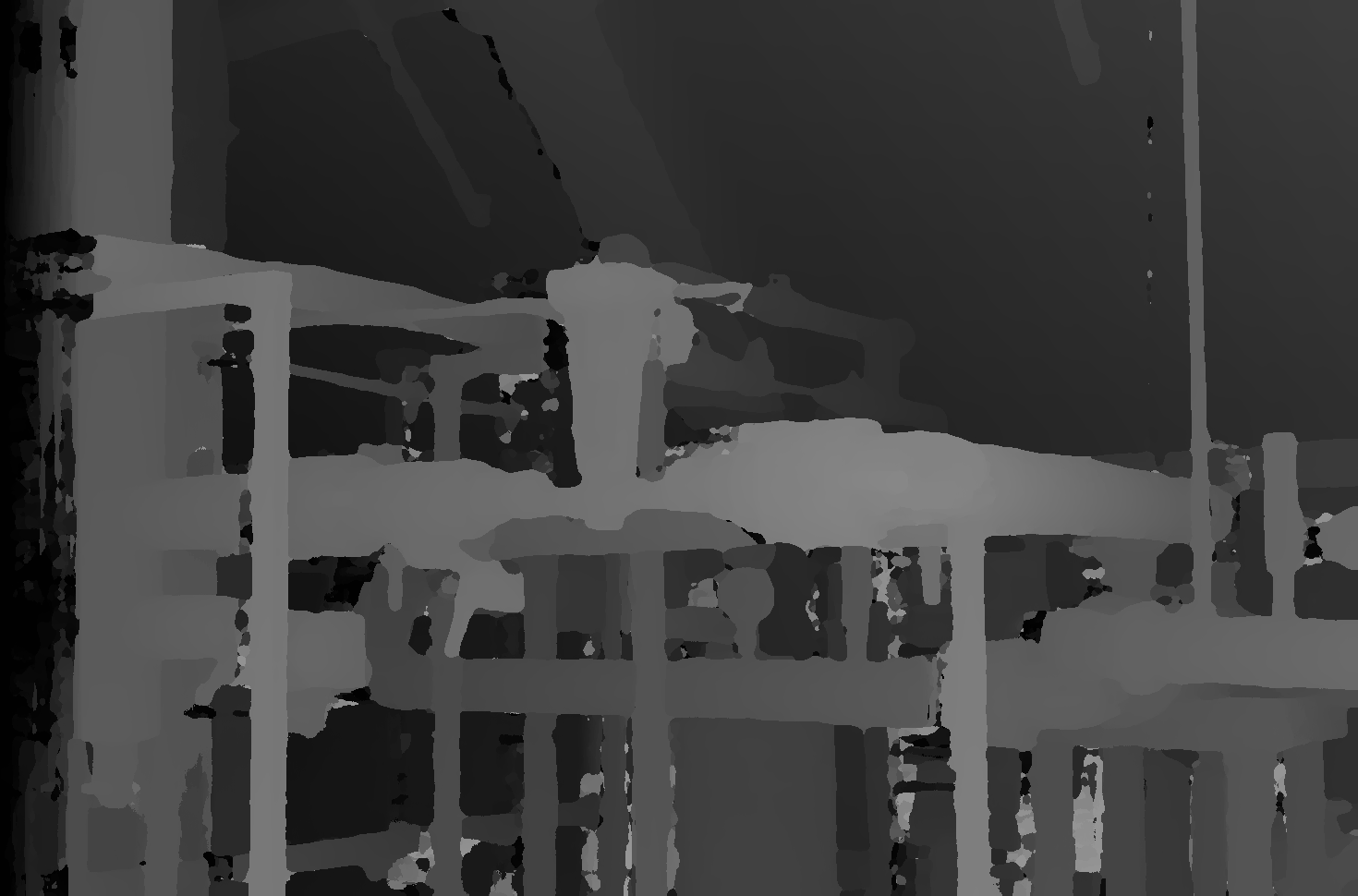}
        \caption{\centering Target image}
    \end{subfigure}%
    ~
        \begin{subfigure}[b]{0.2\textwidth}
        \centering
        \includegraphics[trim={25cm 0cm 10cm 20cm},clip,width=\textwidth]{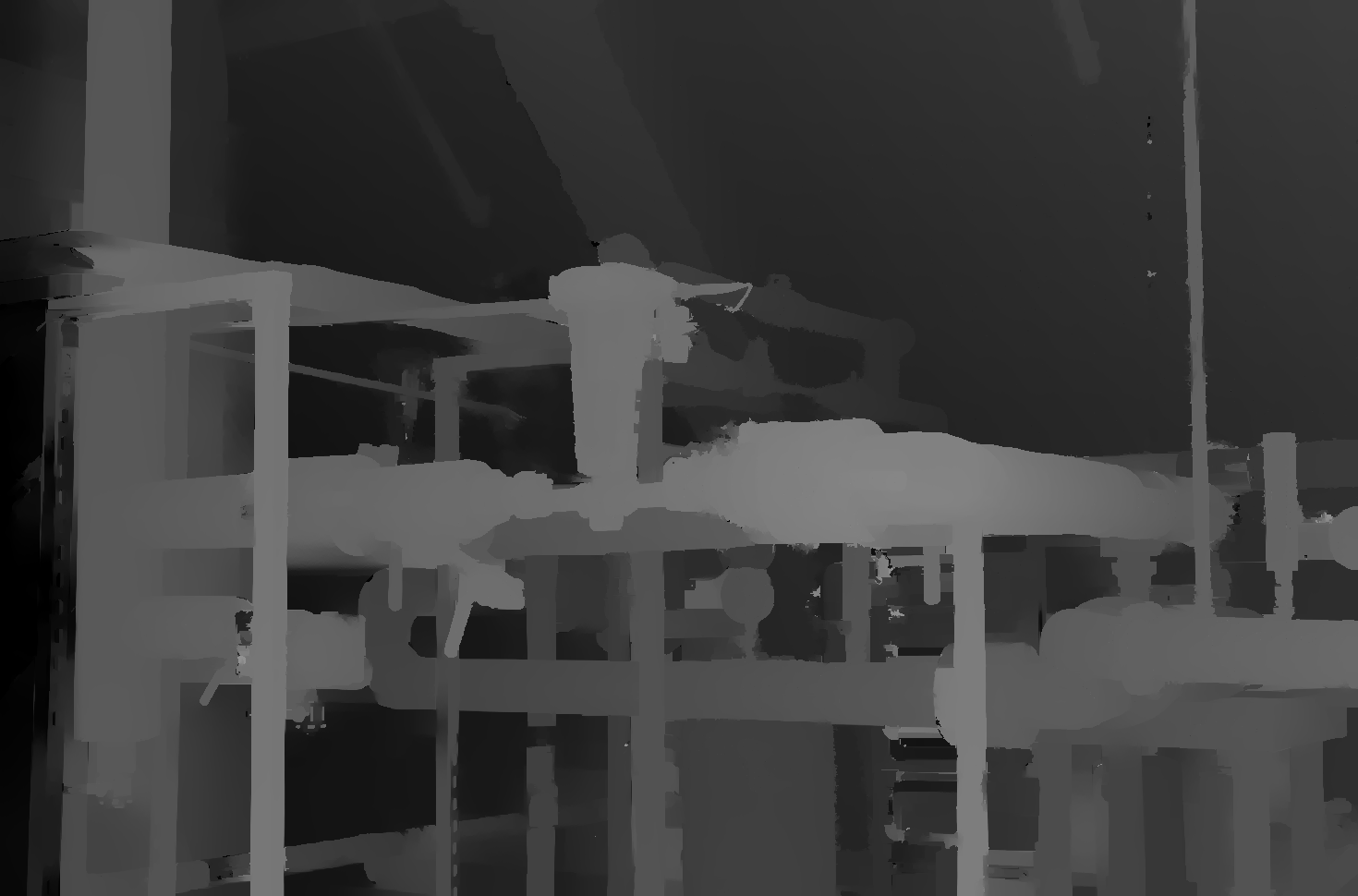}
        \caption{\centering Our result}
    \end{subfigure}%
     \caption{\label{fig:stereo_pipes_closeup}Example stereo optimization for a closeup for of the Middlebury Pipes scene. (a) Color image. (b) Ground-truth disparity. (c) Target obtained using MC-CNN~\cite{zbontar2016stereo}. (d) Our result.}
\end{figure}

\begin{figure}
  \centering
    \begin{subfigure}[t]{0.28\textwidth}
        \centering
        \includegraphics[width=\textwidth] {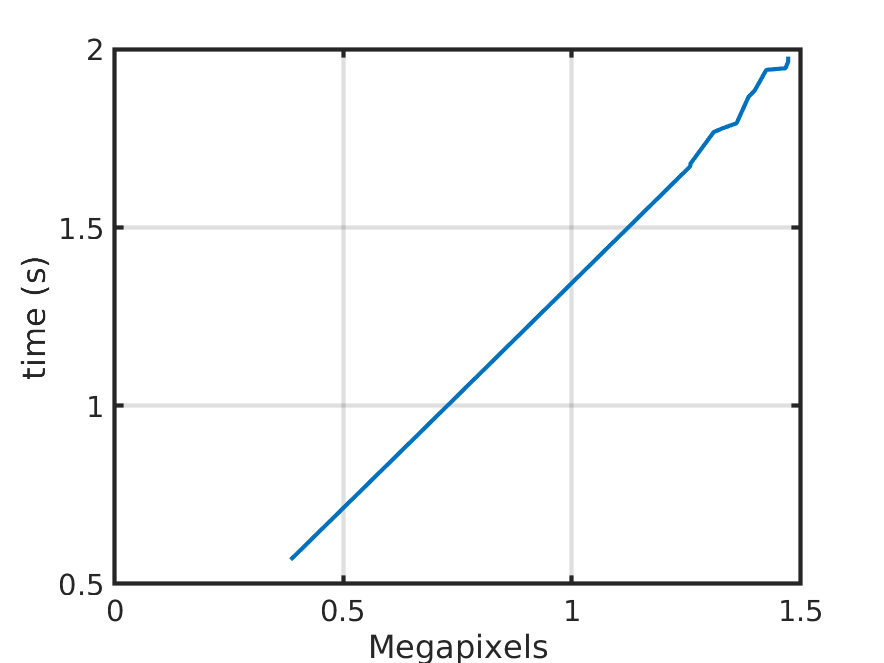}
        \caption{\centering Time vs. megapixels}
    \end{subfigure}%
~
  \centering
    \begin{subfigure}[t]{0.28\textwidth}
        \centering
        \includegraphics[width=\textwidth] {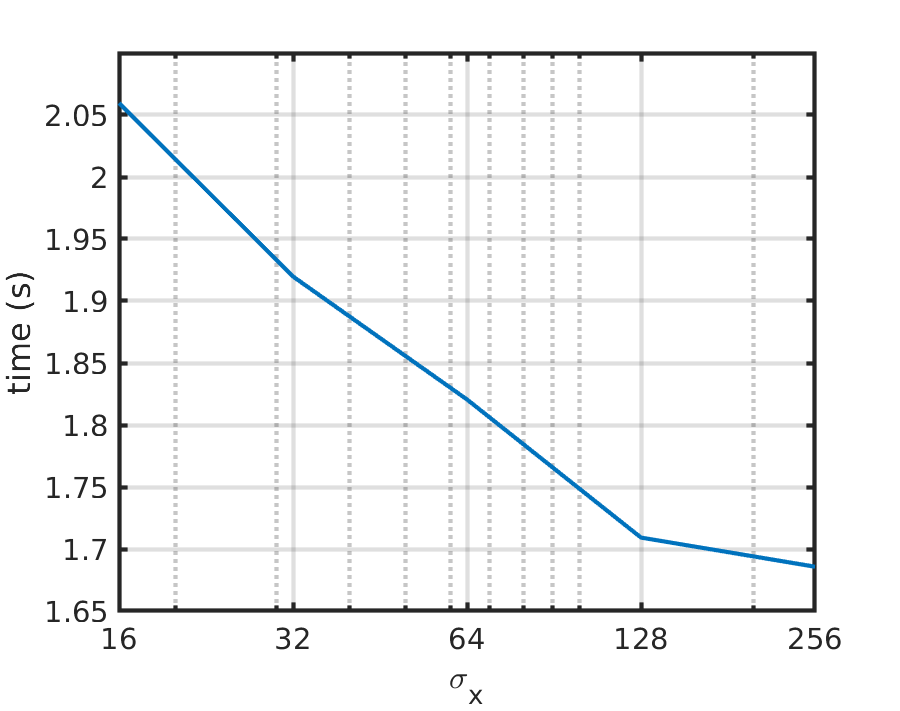}
        \caption{\centering Time vs.~$\sigma_x$}
    \end{subfigure}%
~ 
       \centering
    \begin{subfigure}[t]{0.28\textwidth}
        \centering
        \includegraphics[width=\textwidth] {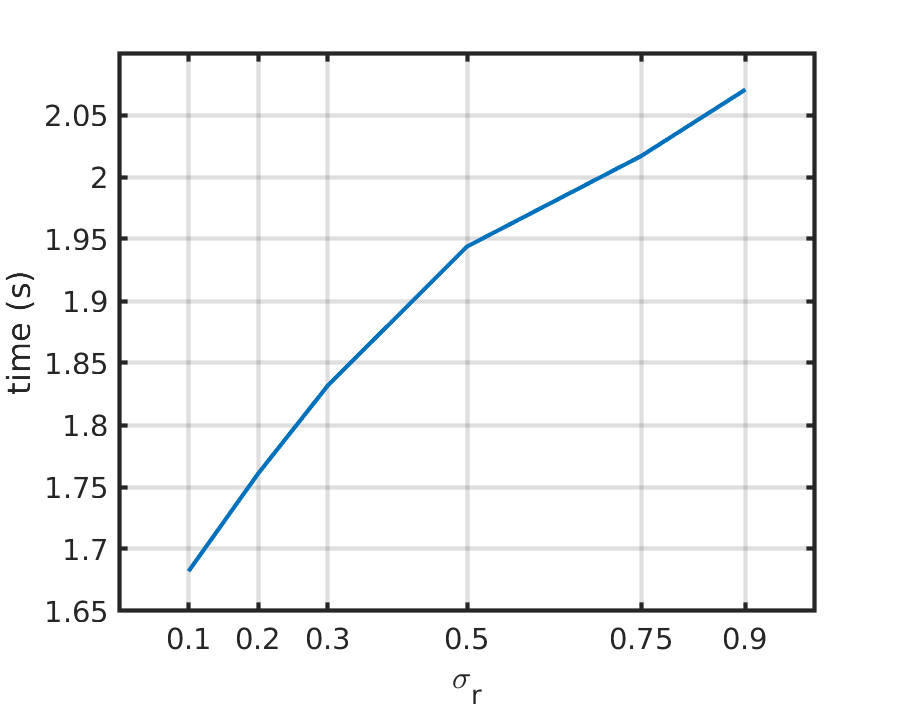}
        \caption{\centering Time vs.~$\sigma_r$}
    \end{subfigure}%
\caption{\label{fig:scale} Our method takes almost the same amount of time even when there is a large difference in blur kernel size, and it linearly scales with image resolution. (a) Time taken for different images in the Middlebury training dataset. (b) Average time taken by DTS for stereo optimization at different spatial blur values. (c) Average time taken by DTS for stereo optimization at different color blur values. All times are at 3000 iterations.}
\end{figure}

\begin{figure}
       \centering
    \begin{subfigure}[t]{0.28\textwidth}
        \centering
        \includegraphics[width=\textwidth] {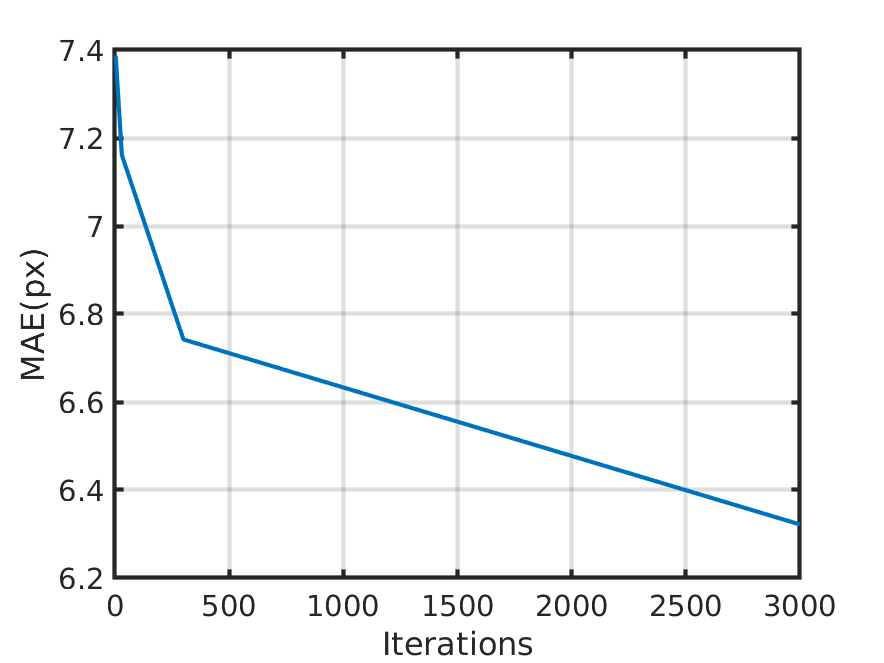}
        \caption{\centering MAE pixel error}
    \end{subfigure}%
    ~ 
       \centering
    \begin{subfigure}[t]{0.28\textwidth}
        \centering
        \includegraphics[width=\textwidth] {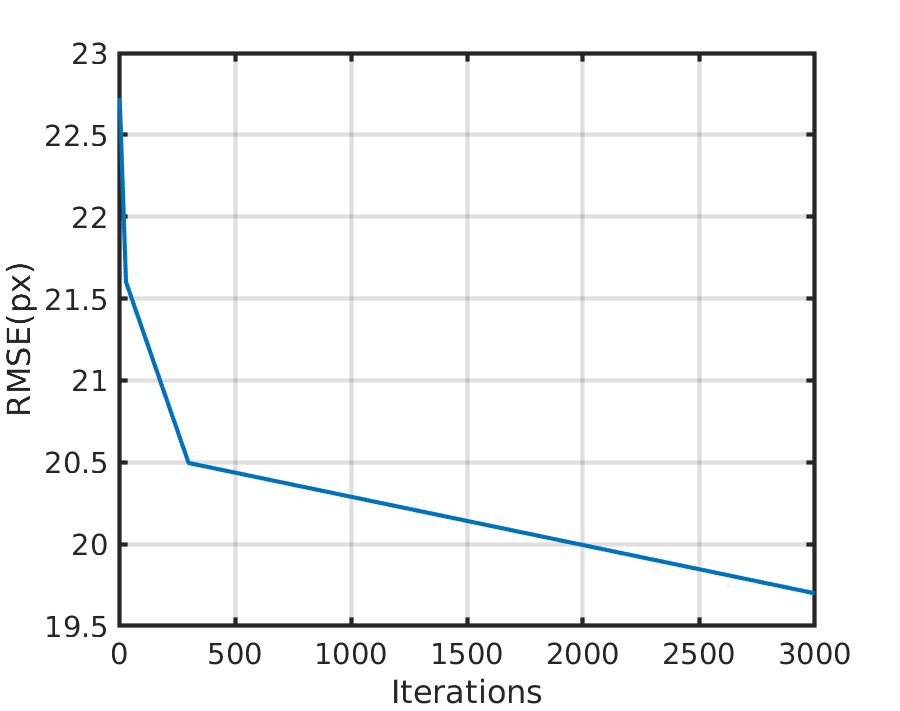}
        \caption{\centering RMSE pixel error}
    \end{subfigure}%
    ~ 
       \centering
    \begin{subfigure}[t]{0.28\textwidth}
        \centering
        \includegraphics[width=\textwidth] {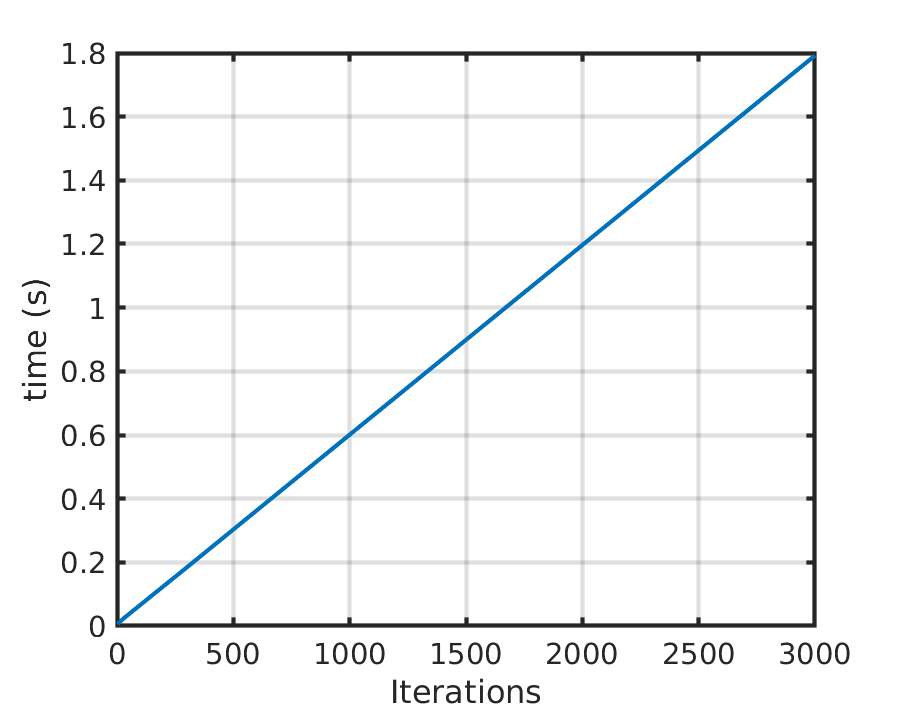}
        \caption{\centering Avg. time(s) vs iterations}
    \end{subfigure}%
\caption{\label{fig:iterations} Dependence of MAE, RMSE and time on the number of iterations. (a-b) The errors reduce quickly for the first 300 iterations and then reduces gradually for further iterations.
The average time taken increases linearly, as shown in (c). This provides a trade-off which can be chosen according to the application.}
\end{figure}

We now present quantitative evaluation of our framework as well as timing performance.
\paragraph{Stereo Optimization}
For the quantitative evaluation of our method, we use the Middlebury dataset~\cite{scharstein2014high}.
Barron and Poole~\cite{barron2016fast} used MC-CNN~\cite{zbontar2016stereo} as their initialization, and for a fair comparison we also use it as our target disparity map. Table~\ref{tab:stereo_middlebury} shows our results for the training set, where we present the mean absolute error (MAE), root mean square error (RMSE), time per megapixel, and time normalized by number of disparity hypotheses for non-occluded regions and for all pixels.
All of these values were determined by the Middlebury evaluation website, and all of our times include the time to calculate MC-CNN on the target disparity maps.
The timing for BL-Solver and our method shows the additional time spent in processing MC-CNN, and the total value in the paranthesis. 
Note that we obtain a huge performance boost compared to MC-CNN at a marginal overhead in time, and we have similar performance with Barron and Poole~\cite{barron2016fast}, especially in non-occluded pixels, while running in only a fraction of their time.
The results for the test set show that our method achieves comparable results in non-occluded regions, while having significant computational savings.
We used $\sigma_{x} = \sigma_{y} = 64px$, $\sigma_{r} = 0.25$ with RGB colors normalized to a range of [0,1], $\lambda = 0.99$, and $\gamma =0.001$.
These parameters were found to work best via grid search strategy on the Middlebury training data.
We ran a gradient descent algorithm for 3000 iterations in this experiment with a step size of 0.99 times the gradient.
Fig.~\ref{fig:stereo_jadeplant_closeup} and Fig.~\ref{fig:stereo_pipes_closeup} show zoomed regions from the Jadeplant and Pipes scene to highlight that we improve the target disparity maps from MC-CNN~\cite{zbontar2016stereo} to estimate sharp depth edges.
\begin{table}
\centering
\begin{tabular}{|c|c|c|c|c|}
\hline 
Training &  &  & & \\ 
\hline 
Algorithm & MAE(px) & RMSE(px) & time/MP(s) & time/GD(s) \\ 
 & no occ$\vert$ all & no occ$\vert$ all &  &  \\ 
\hline 
MC-CNN~\cite{zbontar2016stereo} & 3.81 $\vert$ 11.8 & 18.0 $\vert$ 36.6 & 83.3 & 259 \\ 
\hline 
MC-CNN + BL-Solver~\cite{barron2016fast} & 2.60 $\vert$ 6.66 & 10.2 $\vert$ 20.9 & 42.7 (126) & 153 (412) \\ 
\hline 
MC-CNN+DTS (ours) & 3.02 $\vert$ 9.12 & 10.8 $\vert$ 27.4 & 5.9 (89.2) & 19 (278) \\ 
\hline 
Testing &  &  & & \\ 
MC-CNN~\cite{zbontar2016stereo} & 3.82$\vert$ 17.9 &    21.3 $\vert$ 55.0 &112&254\\ 
\hline 
MC-CNN + BL-Solver~\cite{barron2016fast}& 2.67$\vert$ 8.19 & 15.0$\vert$ 29.9& 28 (140)& 91 (345)\\ 
\hline 
MC-CNN+DTS (ours) &  3.78$\vert$ 14.6 & 17.6$\vert$  43.4&10 (122)& 23 (277)\\ 
\hline 
\hline 
\end{tabular} 
\caption{\label{tab:stereo_middlebury}Performance comparison on training images from the Middlebury dataset. The timing for BL-Solver and our method shows the additional time spent in processing MC-CNN, and the total value in the paranthesis. Our method takes a fraction of time as compared to Barron and Poole~\cite{barron2016fast} to obtain a significant reduction in error versus MC-CNN.}
\end{table}
\paragraph{Depth super-resolution}
We use the dataset introduced by Ferstl~\etal~\cite{ferstl2013image}to evaluate our method for depth super-resolution.
This dataset consists of three scenes (Art, Books, and Moebius) with added noise at 2, 4, 8, and 16x levels of upsampling.
We used $\sigma_{x} = \sigma_{y} = 20\times2^{f}px$ where $f$ is the amount of upsampling.
We used $\sigma_{r} = 0.25$ with RGB colors normalized to a range of [0,1], $\lambda = 0.99$, and 10 iterations of the gradient descent with a step size of 0.99.
In Table~\ref{tab:depth_super_resolution}, we present the RMS and mean geometric errors for each scene.
Data for the bicubic and BL-Solver were produced by using the data and code provided by Barron~\etal~\cite{jon_barron_website}.
We also used the same code to evaluate our method.
Our method and the BL-Solver used the bicubic upsampling as the target image.
Our method is 10x times faster than Barron~\etal~\cite{barron2016fast} while achieving comparable performance on most images, especially images which have higher upsampling factors. 
Our time is the average over all images and includes 0.007 seconds required for bicubic upsampling.
\begin{table}
\centering
\begin{tabularx}{\textwidth}{|c|c|c|c|c|c|c|c|c|c|c|c|c|c|c|}
\hline
\multirow{2}{*}{} &
      \multicolumn{4}{c}{Art} &
      \multicolumn{4}{c}{Books} &
      \multicolumn{4}{c|}{Moebius} & Avg. & Time\\ 
   & 2x&  4x & 8x  &16x &
    2x & 4x & 8x  &16x &
    2x & 4x & 8x  &16x &  (px) & (s) \\ 
\hline 
Bicubic & 5.32 & 6.07 & 7.27 & 9.59  & 5.00&  5.15 & 5.45 & 5.97 & 5.34 & 5.51 & 5.68 & 6.11& 5.94 & 0.007 \\ 
\hline 
BL-Solver~\cite{barron2016fast} &3.02 &3.91 &5.14 &7.47   &1.41 &1.86 &2.42 &3.34   &1.39 &1.82 &2.40 &3.26 &2.75 &0.234 \\ 
\hline 
DTS (Ours)  &4.58  &5.11 & 5.81  &7.69   & 1.94 & 2.34  &2.85  &3.74   & 1.97  &2.34  &2.89  &3.89 &3.43 &  0.0215  \\ 
\hline 
\end{tabularx}
\caption{\label{tab:depth_super_resolution}Performance of DTS on depth super resolution task. Our method is 10x faster than BL-Solver while having comparable performance in most images, especially images with higher upsampling factors.} 
\end{table}
\paragraph{Scale}
Now we present how our method scales with increasing image resolution and increasing blur kernel sizes.
Our method scales linearly with the number of pixels in the image.
Fig.~\ref{fig:scale}(a) shows the dependence of time in seconds on the number of pixels in the image.
We use the training images from the Middlebury dataset and only show the time consumed by DTS for the stereo task at 3000 iterations.
\vspace{-3mm}
\paragraph{Timing vs blur kernels}
The time taken by our method remains mostly constant in comparison to the vastly different blur kernel sizes.
Fig.~\ref{fig:scale}(b) shows average time taken when we change $\sigma_x = \sigma_y = {16, 32, 128, 256}$, while keeping $\sigma_r =0.25$ constant. 
Fig.~\ref{fig:scale}(c) shows average time taken at different values of  $\sigma_r = {0.1,0.2,0.3,0.5,0.75,0.9}$ with $\sigma_x = \sigma_y = 64$ constant.
All the times are at 3000 iterations.
In fact, there is a weak dependence of time on the blur kernel sizes.
The range of kernel sizes is large, hence the small time changes are negligible in practice.
\paragraph{Iterations}
The number of iterations in the gradient descent scheme affects the accuracy.
Here, we study this effect for the training images of the Middlebury dataset.
Fig.~\ref{fig:iterations} (a) and (b) show the MAE and RMSE, which we calculated for the training set for all the pixels including occlusions in all the training images in the dataset.
The 3000 iterations result is the same as the shown in Table~\ref{tab:stereo_middlebury}, but the numbers are different because the Middlebury evaluation internally uses a weighting scheme, which is not used here.
Both MAE and RMSE reduce very quickly at approximately 300 iterations, and after that the gains are smaller.
The time taken for iterations is linear -- see Fig.~\ref{fig:iterations}(c).
This allows us to easily trade off resolution, quality, and run time depending on the application.

%% file: conclusion.tex
\vspace{-7mm}
\section{Conclusion}
\label{sec:conclusion}
We have presented a novel edge-aware solver that achieves scalable performance across a variety of applicable tasks.
Our method is faster by an order of magnitude compared to the state of the art while performing at comparable accuracy.
The approach is highly parallelizable and scales well w.r.t image resolution, and unlike existing methods, it is independent of blurring kernel size.
A future step is to extend our approach to multi-GPU setting, as well as use advanced optimization methods like conjugate gradient descent to obtain faster convergence.